\documentclass[12pt,a4paper]{article}
\usepackage[margin=2.5cm]{geometry}
\usepackage{authblk}
\usepackage{graphicx}
\usepackage{multicol,multirow}
\usepackage{amsmath,amssymb,amsfonts}
\usepackage{mathrsfs}
\usepackage{amsthm}
\usepackage{bbm}
\usepackage{rotating}
\usepackage{appendix}
\usepackage{array,booktabs,xcolor}   
\usepackage{caption}
\usepackage{natbib}                  
\usepackage[colorlinks,allcolors=blue]{hyperref}

\usepackage{algorithm}
\usepackage{algpseudocode}

\algrenewcommand\algorithmicrequire{\textbf{Input:}}
\algrenewcommand\algorithmicensure{\textbf{Output:}}

\newtheorem{theorem}{Theorem}[section]
\newtheorem{lemma}[theorem]{Lemma}
\theoremstyle{definition}
\newtheorem{remark}[theorem]{Remark}

\numberwithin{equation}{section}

\begin{document}

\title{Federated Learning for the Design of Parametric\\
Insurance Indices under Heterogeneous Renewable\\
Production Losses}

\author[1]{Fallou Niakh}
\affil[1]{Laboratory in Finance and Insurance (LFA),
CREST -- Center for Research in Economics and Statistics,
ENSAE IP Paris, Palaiseau, France\\
\texttt{fallou.niakh@ensae.fr}.}

\date{\today}

\maketitle

\begin{abstract}
We propose a federated learning framework for the calibration of parametric insurance indices under heterogeneous renewable energy production losses. Producers locally model their losses using Tweedie generalized linear models and private data, while a common index is learned through federated optimization without sharing raw observations. The approach accommodates heterogeneity in variance and link functions and directly minimizes a global deviance objective in a distributed setting. We establish theoretical guarantees under sub-exponential covariate distributions, showing that the Lipschitz constants of the local Tweedie objectives scale as $\frac{1}{\phi_i}$, where $\phi_i$ is the dispersion parameter of producer $i$. This heterogeneity in smoothness causes naive federated averaging to be biased toward producers with stable microclimates --- precisely those least in need of basis-risk protection --- and motivates the use of corrected aggregation schemes. We implement and compare FedAvg, FedProx and FedOpt, and benchmark them against an existing approximation-based aggregation method. A progressive pool expansion experiment involving up to 121 solar farms in Germany reveals that the approximation-based method becomes entirely non-computable beyond the pool of 50 farms, while federated learning remains valid and actively improves as the pool grows. Federated learning is also over $250\times$ faster than the approximation-based benchmark, establishing it as the only computationally and statistically valid approach for heterogeneous producer pools.
\end{abstract}



\newpage

\section{Introduction}

The rapid expansion of renewable energy production has intensified the exposure of energy producers and investors to weather-related risks. Solar and wind power generation are inherently dependent on meteorological conditions, leading to substantial revenue volatility driven by forecast errors and localized climate variability. In this context, parametric insurance has emerged as a promising risk transfer instrument, offering fast payouts, transparency, and reduced transaction costs by linking indemnities to observable weather indices rather than realized losses. At the same time, the financial losses incurred by individual producers are commercially sensitive data that cannot be freely shared across competitors, creating a fundamental tension between the need for collaborative index calibration and the imperative of data confidentiality.

Despite these advantages, parametric insurance contracts suffer from a well-documented limitation: \emph{basis risk}. Basis risk arises when the chosen index fails to accurately reflect the actual financial losses experienced by insured agents. This issue is particularly acute in renewable energy markets, where production sites are geographically dispersed and subject to heterogeneous microclimatic conditions. Designing indices that are both robust and representative across multiple producers, therefore, remains a central challenge.

In recent work \citep{Niakh2024ANOR}, we introduced a peer-to-peer (P2P) framework for mitigating basis risk in renewable production parametric insurance, based on an approximation-based index calibration method hereafter referred to as \emph{ANOR method}. The proposed mechanism combined a parametric insurance layer with an ex-post redistribution scheme, where residual discrepancies between payouts and realized losses were shared among participants. A key component of this framework was the construction of a common meteorological index obtained by aggregating individual producer sensitivities to weather covariates.

Methodologically, the aggregation relied on a second-order Taylor approximation of the conditional expectation of individual losses, leading to a tractable optimization problem under a common coefficient vector. While this approach yielded strong empirical performance and interpretability, its theoretical validity was contingent on a restrictive assumption: individual regression coefficients linking weather covariates to financial losses must not be excessively heterogeneous. In practice, this assumption may be violated due to differences in geographical location, technology, operational constraints, and exposure profiles across producers.

These limitations motivate the exploration of alternative aggregation paradigms capable of handling structural heterogeneity without relying on analytical approximations.

Federated learning (FL) has emerged as a powerful framework for training statistical and machine learning models across decentralized data sources while preserving data locality and privacy \citep{McMahan2017}. In contrast to centralized learning, FL operates by iteratively exchanging model updates rather than raw data between clients and a coordinating server. This paradigm is particularly well suited for regulated industries such as insurance and finance, where data sharing is constrained by privacy, confidentiality, and governance requirements. In the renewable energy sector specifically, production and financial loss data are subject to strong competitive confidentiality constraints, as they reflect each producer's contract position, hedging strategy, and exposure to imbalance penalties --- information that producers are typically unwilling to disclose to competitors or to a centralized data processor.

Federated learning can be categorized according to the structure of the data across clients. In \emph{horizontal federated learning}, clients share the same feature space but observe different samples, while in \emph{vertical federated learning}, clients observe the same individuals but possess complementary feature sets. The problem studied in this paper naturally falls within the horizontal federated learning framework: each renewable energy producer observes its own historical losses and exposures, described by a common set of meteorological covariates, but over distinct realizations and local conditions.

Several federated learning strategies have been proposed in the literature. The most widely used approach, FedAvg \citep{McMahan2017}, aggregates client-specific model parameters via weighted averaging. Extensions such as FedProx \citep{Li2020FedProx} introduce proximal regularization to stabilize training under data heterogeneity, while adaptive methods like FedOpt \citep{Reddi2021FedOpt} leverage server-side optimization schemes to improve convergence and robustness.

In recent years, FL has gained increasing attention in actuarial science and insurance analytics. Applications include claims frequency prediction, fraud detection, pricing, and risk classification, where sensitive policyholder data cannot be centralized. Notable contributions include federated approaches to life insurance risk prediction \citep{gupta2022risk}, motor insurance claims modeling \citep{yin2024federated}, and health insurance analytics \citep{Rieke2020FLHealth}. These studies highlight FL's ability to reconcile predictive performance with regulatory and ethical constraints.

However, existing insurance applications of federated learning have predominantly focused on classification or regression tasks aimed at individual risk prediction. To the best of our knowledge, federated learning has not yet been systematically explored as a tool for \emph{parametric index design} and collective risk-sharing mechanisms.

\paragraph{Main Contribution}

This paper proposes a federated learning (FL) framework for parametric index design under heterogeneous generalized linear models (GLMs). Each producer is exposed to a random financial loss modeled through a Tweedie GLM, the observations of which are commercially sensitive and cannot be shared with other producers, allowing for producer-specific dispersion parameters and non linear link functions. Rather than enforcing homogeneity through analytical approximations or centralized estimation, we reformulate index calibration as a distributed optimization problem and leverage federated learning to estimate a shared, transparent index structure based on observable meteorological covariates.

The primary contribution of this work is conceptual and methodological: we show that federated learning provides a principled and scalable aggregation mechanism for parametric insurance index design when local loss-generating processes are structurally heterogeneous and data cannot be centralized. The proposed framework accommodates multiple sources of heterogeneity --- including differences in dispersion parameters, link function exponents, and local data distributions --- while preserving the linear and interpretable form required for contractibility and regulatory acceptance of parametric insurance products.

From a methodological standpoint, we adapt and compare several standard federated optimization schemes --- FedAvg, FedProx, and FedOpt --- within a Tweedie GLM setting characterized by heavy-tailed losses and producer-level heterogeneity. We establish theoretical guarantees under sub-exponential covariate distributions, showing that the Lipschitz constants of the local Tweedie objectives scale as $\frac{1}{\phi_i}$, where $\phi_i$ is the dispersion parameter of producer $i$. This result implies that naive federated averaging is biased toward producers with stable microclimates --- precisely those least in need of basis-risk protection --- and provides a principled motivation for the corrected aggregation schemes FedProx and FedOpt. This contribution is important because federated learning is most often studied in high-dimensional or deep learning contexts, whereas here it is embedded in a classical actuarial modeling framework. We analyze the convergence behavior, stability, and uncertainty of these
algorithms under stochastic local updates, highlighting the trade-offs between simplicity, regularization, and adaptive server-side optimization.

The empirical contribution of this work is twofold. To ensure a clean and controlled comparison with existing aggregation-based approaches, we apply the proposed framework to the same empirical dataset used in \cite{Niakh2024ANOR}. Within the original pool of 50 solar farms, FedOpt converges faster and to a lower global deviance than FedAvg and FedProx, while all three methods yield coefficient vectors below the ANOR benchmark, consistently with the fact that FL and ANOR minimize different objectives. Beyond this regime, a progressive pool expansion experiment involving up to 121 solar farms reveals that the approximation-based method becomes entirely non-computable as the pool grows, while federated learning not only remains
valid but actively improves: the global Tweedie deviance decreases from $2.23$ at $k = 50$ to $1.74$ at $k = 121$. Federated learning is over $250\times$ faster than the approximation-based benchmark and is the only valid approach for heterogeneous producer pools.

More broadly, this work contributes to the intersection of actuarial science, climate risk management, and distributed machine learning. It positions federated learning not as a replacement for classical actuarial models, but as a unifying and robust foundation for collaborative parametric index design in settings where data governance, confidentiality constraints, and producer heterogeneity preclude centralized estimation.

\section{Model and Problem Formulation}
\label{sec:model}

\subsection{Economic Setting and Notation}

We consider a group of $N$ renewable energy producers exposed to weather-related production risk. For each producer $i \in \{1,\dots,N\}$, let $X_i \ge 0$ denote the financial loss incurred over a fixed time horizon due to deviations between forecasted and realized production.

Let $\boldsymbol{Y} = (Y_1,\dots,Y_J)^\top$ be a vector of observable meteorological covariates (e.g.\ solar irradiance, temperature, cloud cover), jointly observed by all producers. These covariates are assumed to be publicly observable and measurable in real time, making them suitable candidates for the construction of a parametric insurance index.

Each producer has access only to its own loss observations $\{(X_i^{(d)}, \boldsymbol{Y}^{(d)})\}_{d}$, which cannot be shared due to confidentiality and competitive constraints.

\subsection{Local Structural Models}

We assume that, conditionally on the covariates $\boldsymbol{Y}$, the loss of producer $i$ follows a generalized linear model with Tweedie variance structure. More precisely, for each producer $i$,
\begin{equation}
\label{eq:local_glm}
\mathbb{E}\left[X_i \mid \boldsymbol{Y}\right]
=
\mu_i(\boldsymbol{Y})
=
\left(a_{i0} + \boldsymbol{a}_i^\top \boldsymbol{Y}\right)^{p_i},
\end{equation}
where:
\begin{itemize}
    \item $\boldsymbol{a}_i = (a_{i1},\dots,a_{iJ})^\top \in \mathbb{R}^J$ is a producer-specific
    sensitivity vector,
    \item $p_i > 0$ parameterizes the link function,
    \item $a_{i0} \in \mathbb{R}$ is an intercept term.
\end{itemize}

The conditional variance is assumed to satisfy
\begin{equation}
\label{eq:tweedie_variance}
\mathrm{Var}\left[X_i \mid \boldsymbol{Y}\right]
=
\phi_i \mu_i(\boldsymbol{Y})^{q_i},
\end{equation}
where $\phi_i > 0$ and $q_i \in (1,2)$ are producer-specific Tweedie parameters.

Importantly, the vectors $(\boldsymbol{a}_i, p_i, q_i, \phi_i)$ are \emph{heterogeneous across producers}. They reflect differences in technology, geographical exposure, and local weather patterns. These parameters are latent at the system level and cannot be estimated centrally.

\subsection{Parametric insurance constraint and common index}

Parametric insurance contracts require the definition of a single observable index shared across all insured agents. This requirement is not a modeling convenience but reflects the operational and legal nature of parametric products. We therefore introduce a \emph{common parametric index} of the form
\begin{equation}
\label{eq:common_index}
Z = \boldsymbol{a}^\top \boldsymbol{Y},
\end{equation}
where $\boldsymbol{a} = (a_1,\dots,a_J)^\top \in \mathbb{R}^{J}$ is a vector of coefficients common to all producers. This index is transparent, objectively verifiable by all parties --- including regulators, reinsurers, and capital market investors --- and immune to manipulation or moral hazard, since it depends solely on publicly observable meteorological quantities rather than on realized individual losses. By contrast, conditioning directly on the full covariate vector $\boldsymbol{Y}$ to compute producer-specific payouts $\mathbb{E}[X_i \mid \boldsymbol{Y}]$ would undermine the key properties that make parametric insurance operationally attractive: automatic and fast settlement, absence of loss adjustment procedures, and suitability for trading and securitization. In practice, parametric products traded on capital markets or structured as catastrophe bonds require a single trigger index common to all insured parties, with individual exposure captured only through simple contract parameters such as coverage limits, deductibles, or notional quantities.

Parametric insurance is triggered when the index $Z$ exceeds a contractual attachment point $z_0 > 0$ beyond which an unanticipated production loss is deemed to have occurred. The parametric contract compensates producer $i$ according to:
\begin{equation}
m_i(z) =
\begin{cases}
\mathbb{E}[X_i \mid Z = z] & \text{if } z > z_0, \\
0 & \text{if } z \leq z_0.
\end{cases}
\label{fctmi}
\end{equation}
In return, the producer pays a premium that accounts for administrative costs, risk margins, and other expenses. When $Z \leq z_0$, no compensation is provided regardless of the realized loss $X_i$ --- this is the defining feature of parametric insurance and the primary source of basis risk for the insured.

Under this trigger structure, producer $i$ approximates its conditional expected loss \emph{on triggered days} by
\begin{equation}
\label{eq:approx_mean}
\tilde{\mu}_i(\boldsymbol{Y}; \boldsymbol{a})
=
Z^{p_i}, \qquad Z > z_0.
\end{equation}
The vector $\boldsymbol{a}$ should \emph{not} be interpreted as a structural parameter of any individual producer. Rather, it represents a collective projection of heterogeneous local models onto a common index, required by the parametric insurance design. For estimation purposes, only observations in the triggered region $\{Z > z_0\}$ enter the federated objective, since the contract pays zero when $Z \leq z_0$ and the approximation quality of $\tilde{\mu}_i$ below the attachment point is therefore contractually irrelevant. This restriction is not an additional modeling assumption but a direct consequence of the contract structure \eqref{fctmi}. It is formalized as Assumption~(A3) in Section~\ref{sec:federated}, where it also ensures the well-posedness of the gradient computations on the triggered region under the light-tailed covariate assumption~(A1).

However, the compensation $m_i(Z)$ does not perfectly match the actual loss, since it depends on weather parameters rather than on realized production data. This discrepancy defines the \emph{basis risk}:
\begin{equation}
\varepsilon_i = X_i - m_i(Z).
\end{equation}
Note that basis risk takes two qualitatively different forms depending on the trigger status. When $Z > z_0$, the producer receives compensation $m_i(Z) = \mathbb{E}[X_i \mid Z]$ but may still suffer a residual loss $\varepsilon_i = X_i - \mathbb{E}[X_i \mid Z]$ due to the imperfect correlation between the index and individual losses. When $Z \leq z_0$, the producer receives no compensation at all, so $\varepsilon_i = X_i$ --- the entire loss is borne by the producer. The design of the index $\boldsymbol{a}$ therefore affects basis risk through two channels: by improving the fit of the compensation function in the triggered region, and by influencing the probability that the trigger is activated for producers who have suffered significant losses.

Beyond these operational considerations, the use of a shared scalar index enables \emph{basis risk pooling} within the group of producers. Since all producers are exposed to the same trigger $Z$, their individual basis risks $\varepsilon_i$ are driven by idiosyncratic factors conditionally on $Z > z_0$. Under standard conditional independence assumptions, the aggregate basis risk $S_\varepsilon = \sum_{i=1}^N \varepsilon_i$ is diversifiable within the pool, and natural risk-sharing rules --- such as allocating aggregate basis risk proportionally to conditional variance \citep{Niakh2024ANOR} --- allow producers to reduce their residual exposure collectively. This diversification argument is structurally meaningful only when all producers share a common trigger, and would be lost if each producer conditioned on a different functional of $\boldsymbol{Y}$.

We aim to design $Z$, equivalently to estimate $\boldsymbol{a}$, using federated learning.

\subsection{Loss Function and Global Objective}

To quantify the discrepancy induced by enforcing a common index on the triggered region, we rely on the Tweedie deviance. For producer $i$, the pointwise loss is defined as
\begin{equation}
\begin{aligned}
d_{p_i, q_i,\phi_i}(x,\mu) = & \frac{2}{\phi_i(1-q_i)(2-q_i)}
x^{2-q_i}
-
\frac{2}{\phi_i(1-q_i)} x\mu^{1-q_i}
\\
& + \frac{2}{\phi_i(2-q_i)} \mu^{2-q_i},
\end{aligned}
\label{eq:tweedie_deviance}
\end{equation}
for $\mu > 0$. This is the building block from which all local and global objectives in the federated framework are constructed, as detailed in Section~\ref{sec:federated}.

The global optimization problem is then given by
\begin{equation}
\label{eq:global_objective}
\min_{\boldsymbol{a} \in \mathcal{A}}
\;
\sum_{i=1}^N \omega_i
\;
\mathbb{E}
\left[
d_{p_i, q_i,\phi_i}
\left(
X_i,
\tilde{\mu}_i(\boldsymbol{Y}; \boldsymbol{a})
\right)
\;\middle|\; Z > z_0
\right],
\end{equation}
where $\omega_i > 0$ are aggregation weights, typically proportional to exposure or portfolio size, and $\mathcal{A} = \{\boldsymbol{a} : \|\boldsymbol{a}\|_2 \leq M\}$ is the feasible set introduced in Assumption~(A2) of Section~\ref{sec:federated}. Here $\omega_i$ is taken as the capacity of the farm.

This problem captures the fundamental trade-off of parametric insurance: a single index must provide an acceptable approximation of heterogeneous loss processes across producers on the days where compensation is contractually due.

\subsection{Why Federated Learning?}
Direct minimization of \eqref{eq:global_objective} is infeasible in practice for two distinct reasons. First, the joint distribution of $(X_i, \boldsymbol{Y})$ is not centrally observable: each producer holds only its own loss history, and these data cannot be pooled. Second, and more fundamentally, the financial loss data $X_i$ are highly sensitive. For solar power producers, $X_i$ reflects the economic consequence of a production shortfall relative to the day-ahead forecast — a quantity that encodes the producer's contract position, hedging strategy, feed-in tariff, and exposure to imbalance penalties. These are commercially confidential data that producers are typically unwilling to share with competitors, with an insurer acting as a central data processor, or with any third party. Centralizing such data would raise serious confidentiality concerns and may conflict with contractual or regulatory constraints on data sharing.\newline
Federated learning resolves this tension by enabling the collaborative optimization of the common coefficient vector $\boldsymbol{a}$ through local gradient updates, without sharing raw observations. Each producer computes gradients locally using its own private loss data and transmits only these parameter updates to the server. The financial loss series $\{X_{i,d}\}$ never leave the producer's premises, and the server learns only the aggregated index weights. This privacy guarantee is not merely a technical nicety: it is a prerequisite for the practical deployability of the framework. Without it, risk-averse producers would have strong incentives to withhold their data or misreport their losses, making any form of collaborative calibration infeasible in practice.\newline
This framework naturally accommodates heterogeneity in $(p_i, q_i, \phi_i)$ and allows for flexible aggregation schemes, which are detailed in Section ~\ref{sec:federated}.

\subsection{The approximation-based approach and its limitations}
\label{sec:anor_recall}

Before introducing the federated learning framework, we briefly recall the index calibration method proposed in \cite{Niakh2024ANOR}, which serves as our benchmark throughout the empirical analysis. This recall is necessary to make explicit the assumption on which that method rests and to introduce the heterogeneity measure used to document its breakdown in Section~\ref{sec:results_expansion}.

\subsubsection*{The Taylor approximation}

The ANOR method seeks to minimize the expected conditional variance of the basis risk across producers. To make this objective tractable, it approximates the conditional expectation $\mathbb{E}[X_i \mid Z = z]$ by a second-order Taylor expansion around the common index $Z =\boldsymbol{a}^\top \boldsymbol{Y}$. Specifically, rewriting the individual linear score as
\begin{equation}
\label{eq:score_decomp}
a_{i0} + \boldsymbol{a}_i^\top \boldsymbol{Y}
\;=\;
a_{i0} + Z + \|\boldsymbol{a}_i - \boldsymbol{a}\|\,
\boldsymbol{\alpha}_i^\top \boldsymbol{Y},
\end{equation}
where $\boldsymbol{\alpha}_i = \frac{(\boldsymbol{a}_i -
\boldsymbol{a})}{\|\boldsymbol{a}_i - \boldsymbol{a}\|}$, the conditional
expectation expands as
\begin{equation}
\label{eq:taylor_expansion}
\mathbb{E}[X_i \mid \boldsymbol{a}_i^\top \boldsymbol{Y} = z]
\;=\;
\mathbb{E}[X_i \mid Z = z]
+ \|\boldsymbol{a}_i - \boldsymbol{a}\|\, L_i(z, \boldsymbol{a})
+ \frac{1}{2}\|\boldsymbol{a}_i - \boldsymbol{a}\|^2
  F_i(z, \boldsymbol{a})
+ o\!\left(\|\boldsymbol{a}_i - \boldsymbol{a}\|^2\right),
\end{equation}
where $L_i$ and $F_i$ are first- and second-order correction terms involving conditional covariances and derivatives of the conditional mean \citep{Niakh2024ANOR}.

\subsubsection*{The small-heterogeneity assumption}

The validity of the approximation \eqref{eq:taylor_expansion} as a basis for optimization requires that the remainder $o(\|\boldsymbol{a}_i - \boldsymbol{a}\|^2)$ be negligible uniformly across producers. This holds if and only if
\begin{equation}
\label{eq:homogeneity_assumption}
\delta_i \;=\; \|\boldsymbol{a}_i - \boldsymbol{a}\|_2
\end{equation}
is small for all $i$. When this condition is satisfied, the conditional variance $\mathrm{Var}[X_i \mid Z = z]$ can be approximated as a function of $\boldsymbol{a}$ alone:

\begin{equation}
    \text{Var}[X_i \mid Z = z] \simeq \phi_i \left(\mathbb{E}[X_i \mid \boldsymbol{a}_i^{\prime} \boldsymbol{Y}=z] - \|\boldsymbol{a}_i - \boldsymbol{a}\| L_i(z,\boldsymbol{a}) 
-\frac{1}{2} \|\boldsymbol{a}_i - \boldsymbol{a}\|^2 F_i(z,\boldsymbol{a})\right)^{q_i}.
    \label{varapprox}
\end{equation}
This yields a tractable optimization problem for calibrating the index $Z$:
\begin{equation}
\label{problem}
\begin{aligned}
&\min_{\boldsymbol{a} \in \mathcal{A}}
\sum_{i=1}^n \mathbb{E}\!\left[
\mathrm{Var}\left[X_i \mid Z\right] \mid Z > z_0
\right] \\
&= \min_{\boldsymbol{a} \in \mathcal{A}}\sum_{i=1}^n \mathbb{E}\!\left[
\phi_i \left(
\mathbb{E}[X_i \mid \boldsymbol{a}_i^{\prime}\boldsymbol{Y}=Z]
- \|\boldsymbol{a}_i - \boldsymbol{a}\|\, L_i(Z,\boldsymbol{a})
- \tfrac{1}{2}\|\boldsymbol{a}_i - \boldsymbol{a}\|^2
F_i(Z,\boldsymbol{a})
\right)^{q_i}
\right].
\end{aligned}
\end{equation}
\subsubsection*{The Taylor remainder as a heterogeneity diagnostic}

To quantify how far a given producer pool is from satisfying \eqref{eq:homogeneity_assumption}, we define the following empirical proxy for the aggregate Taylor remainder at pool size $k$:
\begin{equation}
\label{eq:Rk}
R_k \;=\; \frac{1}{k} \sum_{i=1}^k \delta_i^2
\;=\; \frac{1}{k} \sum_{i=1}^k
\|\boldsymbol{a}_i - \boldsymbol{a}^{ref}\|_2^2,
\end{equation}
where $\boldsymbol{a}^{ref}$ is estimated as the mean of
the individual GLM coefficients $\boldsymbol{a}_i$ across the original 50 farms. When $R_k$ is small, the ANOR approximation is reliable and both methods are expected to produce comparable indices. When $R_k$ grows, the remainder term in \eqref{eq:taylor_expansion} becomes non-negligible, and the ANOR optimization returns numerically invalid results. Section~\ref{sec:results_expansion} tracks $R_k$ as the pool size increases from the original $k = 50$ farms to the full pool of $k = 121$ farms and documents precisely the threshold beyond which the ANOR method collapses.

Federated learning, by contrast, directly minimizes the global Tweedie deviance \eqref{eq:fed_objective} without relying on any Taylor expansion, and therefore remains valid regardless of the magnitude of $\delta_i$. This structural difference — not a marginal improvement in performance — constitutes the primary empirical motivation for the federated learning framework proposed in this paper.

\section{Federated Optimization for Parametric Index Design}
\label{sec:federated}

\subsection{Federated learning formulation}

We aim to solve the global optimization problem introduced in \eqref{eq:global_objective}, namely
\[
\min_{\boldsymbol{a} \in \mathcal{A}}
\sum_{i=1}^N \omega_i
\mathbb{E}
\left[
d_{p_i, q_i,\phi_i}
\left(
X_i,
\tilde{\mu}_i(\boldsymbol{Y};\boldsymbol{a})
\right)
\;\middle|\; Z > z_0
\right],
\]
under the constraint that each producer $i$ observes only its own data. The expectation is taken conditionally on the triggered region $\{Z > z_0\}$, since the parametric contract pays zero when $Z \leq z_0$ and only triggered observations are relevant for calibrating the compensation function $m_i$.

We restrict optimization to the feasible set $\mathcal{A} = \{\boldsymbol{a} : \|\boldsymbol{a}\|_2 \leq M\}$, for some $M > 0$, under the following regularity conditions:

\begin{itemize}
    \item[(A1)] \emph{Light-tailed weather:} the covariate vector $\boldsymbol{Y}$ is sub-exponential, i.e.\ there exist constants $\sigma, b > 0$ such that
    \begin{equation*}
        \mathbb{P}\left(\|\boldsymbol{Y}\|_2 > t\right)
        \;\leq\; 2\exp(-t/b),
        \qquad \text{for all } t \geq \sigma.
    \end{equation*}
    \emph{(All polynomial moments of the weather covariates are finite. Extreme meteorological values are permitted --- in contrast to a bounded-support assumption --- but occur with exponentially decaying probability, consistent with standard modelling of standardized irradiance forecast errors after monthly stationarization.)}

    \item[(A2)] \emph{Bounded index coefficients:} optimization is restricted to $\mathcal{A} = \{\boldsymbol{a} : \|\boldsymbol{a}\|_2 \leq M\}$. \emph{(The index cannot generate unbounded payouts, a standard actuarial restriction \citep{McCullaghNelder1989}.)}

    \item[(A3)] \emph{Strictly positive index on the triggered region:} $\boldsymbol{a}^\top \boldsymbol{Y} \geq z_0 > 0$ almost surely on $\mathcal{D}_i^+ = \{d : \boldsymbol{a}^\top \boldsymbol{Y}^{(d)} > z_0\}$. \emph{(On the days where the parametric contract is triggered, the index is strictly positive by definition of the triggered region, ensuring that $\tilde{\mu}_i = Z^{p_i}$ is well-defined and that all negative-power terms appearing in the Tweedie deviance and its derivatives --- such as $\tilde{\mu}_i^{-q_i}$ and $Z^{p_i - 1}$ --- are bounded away from zero. No assumption is placed on the behaviour of $\boldsymbol{Y}$ below $z_0$, since those observations do not enter the federated objective.)}
\end{itemize}

Assumptions (A1) and (A3) together replace the bounded-covariate assumption commonly used in the federated optimization literature. A bounded-support assumption on $\boldsymbol{Y}$ would be inappropriate in the present context, where extreme weather realizations are precisely the events that parametric insurance is designed to cover. Under (A1), such extremes are permitted with exponentially decaying probability; under (A3), the optimization is restricted to the economically relevant triggered region $\{Z > z_0\}$, on which the index is strictly positive by construction. This is not an additional restriction on the model but a direct consequence of the contract structure: the parametric contract pays zero when $Z \leq z_0$, so those observations carry no information for calibrating the compensation function $m_i$ \eqref{fctmi} and are naturally excluded from the objective.

The shared-index model $\tilde{\mu}_i = Z^{p_i}$ is a deliberate \emph{misspecification} of producer $i$'s true model $\mu_i = (a_{i0} + \boldsymbol{a}_i^\top \boldsymbol{Y})^{p_i}$ from Eq.~\eqref{eq:local_glm}: federated learning forces all producers to share a common $\boldsymbol{a}$ instead of their private $\boldsymbol{a}_i$. Consequently, treating the shared-index Tweedie expression as a fully specified likelihood would be theoretically inconsistent. We therefore adopt the \emph{quasi-likelihood} framework of \cite{Wedderburn1974, McCullaghNelder1989}, a standard tool in actuarial GLMs that requires only specifying the working mean and variance functions, without committing to a complete probability distribution. The working assumptions are:
\begin{equation}
\label{eq:working_moments}
    \mathbb{E}^*[X_i \mid \boldsymbol{Y}] = \tilde{\mu}_i,
    \qquad
    \mathrm{Var}^*[X_i \mid \boldsymbol{Y}] = \phi_i\, \tilde{\mu}_i^{q_i},
\end{equation}
where the asterisks signal that these are working assumptions under the shared-index constraint, not claims about the true data-generating process.

\paragraph{From pointwise deviance to local objectives.}

The pointwise Tweedie deviance $d_{p_i,q_i,\phi_i}$ introduced in Eq.~\eqref{eq:tweedie_deviance} serves as the common building block for all local and global objectives in the federated framework. Three nested objects are constructed from it, each playing a distinct role.

\textbf{Population objective.} The theoretical local objective of producer $i$ is the expected deviance under the shared-index approximation $\tilde{\mu}_i = Z^{p_i}$, restricted to triggered days:
\begin{equation}
\label{eq:loglik}
    \ell_i(\boldsymbol{a})
    \;=\;
    \mathbb{E}\!\left[
    d_{p_i,q_i,\phi_i}\!\left(X_i,\,
    \tilde{\mu}_i(\boldsymbol{Y};\boldsymbol{a})\right)
    \;\middle|\; Z > z_0
    \right]
    \;=\;
    \mathbb{E}\!\left[
    \frac{1}{\phi_i}\!\left(
    \frac{\tilde{\mu}_i^{2-q_i}}{2-q_i}
    - \frac{X_i\,\tilde{\mu}_i^{1-q_i}}{1-q_i}
    \right)
    \;\middle|\; Z > z_0
    \right] + C_i,
\end{equation}
where $C_i$ collects terms depending only on $X_i$ and not on $\boldsymbol{a}$, and therefore plays no role in optimization. The equality between the two expressions follows directly from Eq.~\eqref{eq:tweedie_deviance}. The population objective $\ell_i$ is used in the theoretical analysis of Section~\ref{sec:theory}, where gradients and curvature are computed with respect to it.

\textbf{Empirical local risk.} Since the distribution of $(X_i, \boldsymbol{Y})$ is unknown, $\ell_i(\boldsymbol{a})$ is approximated by its sample counterpart over the triggered subset of the local dataset:
\begin{equation}
\label{eq:local_objective}
F_i(\boldsymbol{a})
\;=\;
\frac{1}{|\mathcal{D}_i^+|}
\sum_{d \in \mathcal{D}_i^+}
d_{p_i,q_i,\phi_i}\!\left(
X_i^{(d)},\,
\tilde{\mu}_i(\boldsymbol{Y}^{(d)};\boldsymbol{a})
\right),
\end{equation}
where
\begin{equation}
\label{eq:triggered_set}
\mathcal{D}_i^+ \;=\; \left\{d \in \mathcal{D}_i \;:\;
\boldsymbol{a}^\top \boldsymbol{Y}^{(d)} > z_0 \right\}
\end{equation}
is the subset of days on which the parametric contract is triggered. By the law of large numbers, $F_i(\boldsymbol{a}) \to \ell_i(\boldsymbol{a})$ as $|\mathcal{D}_i^+| \to \infty$, so minimizing $F_i$ is consistent with minimizing $\ell_i$. This is the object computed by each producer during local training in Algorithm~\ref{alg:federated_tweedie}, and whose gradient $\nabla F_i$ drives the local update steps.

\begin{remark}
The triggered set $\mathcal{D}_i^+$ depends on $\boldsymbol{a}$, which makes the objective $F_i$ non-standard: as $\boldsymbol{a}$ changes during optimization, the set of observations entering the sum may change. In practice, the attachment point $z_0$ is estimated once on a preliminary linear score $\tilde{Z}$ as described in Section 3.2 of \cite{Niakh2024ANOR}. The dataset is filtered before federated learning begins. The triggered set is therefore fixed throughout optimization, and $F_i$ is a standard empirical risk on this filtered dataset.
\end{remark}

\textbf{Global federated objective.} The server seeks to minimize the weighted aggregate of local empirical risks:
\begin{equation}
\label{eq:fed_objective}
F(\boldsymbol{a})
\;=\;
\sum_{i=1}^N \omega_i\, F_i(\boldsymbol{a}),
\end{equation}
which approximates the population-level global objective \eqref{eq:global_objective} conditionally on the triggered region. The gradient dissimilarity and smoothness results of Section~\ref{sec:theory} are stated for the population objects $\ell_i$, and carry over to $F_i$ as $|\mathcal{D}_i^+| \to \infty$.

\subsection{Federated optimization protocol}

The federated optimization proceeds in synchronous rounds. At round $t$, the server broadcasts the current global parameter vector $\boldsymbol{a}^{(t)}$ to all producers. Each producer performs several local optimization steps minimizing its empirical risk $F_i$, starting from $\boldsymbol{a}^{(t)}$, and returns an updated parameter vector $\boldsymbol{a}_i^{(t+1)}$ to the server. The server then aggregates the local updates to form the next global iterate $\boldsymbol{a}^{(t+1)}$. This procedure is summarized in Algorithm~\ref{alg:federated_tweedie}.

\begin{algorithm}[H]
\caption{Federated Learning for Parametric Index Design}
\label{alg:federated_tweedie}
\begin{algorithmic}[1]
\Require Number of rounds $T$, local epochs $E$, step size $\eta$
\Ensure Global parameter $\boldsymbol{a}^{(T)}$

\State Server initializes global parameters $\boldsymbol{a}^{(0)}$

\For{$t = 0$ \textbf{to} $T-1$}
    \State Server broadcasts $\boldsymbol{a}^{(t)}$ to all producers
    \ForAll{producers $i = 1,\dots,N$ \textbf{in parallel}}
        \State $\boldsymbol{a}_i^{(t,0)} \gets \boldsymbol{a}^{(t)}$
        \For{$e = 1$ \textbf{to} $E$}
            \State Sample mini-batch $B_i$
            \State
            $
            \boldsymbol{a}_i^{(t,e)}
            \gets
            \boldsymbol{a}_i^{(t,e-1)}
            -
            \eta \nabla F_i(\boldsymbol{a}_i^{(t,e-1)}; B_i)
            $
        \EndFor
        \State Producer sends $\boldsymbol{a}_i^{(t+1)}
        \gets \boldsymbol{a}_i^{(t,E)}$ to server
    \EndFor
    \State Server aggregates
    $$
    \boldsymbol{a}^{(t+1)}
    \gets
    \mathrm{Aggregate}\left(\{\boldsymbol{a}_i^{(t+1)}\}_{i=1}^N\right)
    $$
\EndFor
\end{algorithmic}
\end{algorithm}

\subsection{Gradient dissimilarity and client drift under Tweedie
heterogeneity}
\label{sec:theory}

Before presenting the specific aggregation rules, we establish the theoretical properties of the population objectives $\ell_i$ under heterogeneous producer parameters. The central message is the following: producers with stable microclimates (small $\phi_i$) generate strong, sharp learning signals, while producers with volatile microclimates (large $\phi_i$) generate weak, noisy signals. Without correction, simple averaging methods systematically over-weight stable producers. This is precisely the wrong outcome from a parametric insurance design standpoint, since stable producers are the least in need of basis-risk protection. The results below make this argument quantitative and motivate the algorithmic choices of Sections~\ref{sec:fedprox} and~\ref{sec:fedopt}.

\subsubsection*{Gradient of the population objective}

By the chain rule applied through $\tilde{\mu}_i = Z^{p_i}$ and $Z = \boldsymbol{a}^\top \boldsymbol{Y}$, the gradient of the population objective $\ell_i$ defined in \eqref{eq:loglik} is:
\begin{equation}
\label{eq:gradient}
    \nabla \ell_i(\boldsymbol{a})
    \;=\; \frac{p_i}{\phi_i}\,
    \mathbb{E}\!\left[
    \frac{\tilde{\mu}_i - X_i}{\tilde{\mu}_i^{q_i}}\,
    Z^{p_i - 1}\, \boldsymbol{Y}
    \;\middle|\; Z > z_0
    \right].
\end{equation}
The corresponding empirical gradient, used in Algorithm~\ref{alg:federated_tweedie}, is obtained by replacing the conditional expectation with a sample average over $\mathcal{D}_i^+$:
\begin{equation}
\label{eq:emp_gradient}
    \nabla F_i(\boldsymbol{a})
    \;=\; \frac{p_i}{\phi_i\, |\mathcal{D}_i^+|}
    \sum_{d \in \mathcal{D}_i^+}
    \frac{\tilde{\mu}_i^{(d)} - X_i^{(d)}}{(\tilde{\mu}_i^{(d)})^{q_i}}\,
    (Z^{(d)})^{p_i - 1}\, \boldsymbol{Y}^{(d)},
\end{equation}
where $\tilde{\mu}_i^{(d)} = (\boldsymbol{a}^\top \boldsymbol{Y}^{(d)})^{p_i}$ and $Z^{(d)} = \boldsymbol{a}^\top \boldsymbol{Y}^{(d)} > z_0$ for all $d \in \mathcal{D}_i^+$ by \eqref{eq:triggered_set}. By the law of large numbers, $\nabla F_i(\boldsymbol{a}) \to \nabla \ell_i(\boldsymbol{a})$ as $|\mathcal{D}_i^+| \to \infty$, so the theoretical properties derived below for $\nabla \ell_i$ apply to $\nabla F_i$ in large samples.

Equations \eqref{eq:gradient}--\eqref{eq:emp_gradient} reveal the first key structural fact: \emph{both population and empirical gradients scale as $\frac{1}{\phi_i}$.} A producer with half the dispersion produces gradients that are twice as large at every parameter value. When the server averages these gradients, producers with stable microclimates dominate, even when all producers carry equal weight $\omega_i$ in the global objective.

\subsubsection*{Curvature of the population objective}

We now examine the local curvature via the quasi-Fisher information matrix $I_i(\boldsymbol{a}) = \mathbb{E}[\nabla^2 \ell_i(\boldsymbol{a}) \mid Z > z_0]$. Differentiating \eqref{eq:gradient} with respect to $\boldsymbol{a}$ yields two terms: a curvature term and a residual term proportional to $X_i - \tilde{\mu}_i$. Under the working mean assumption \eqref{eq:working_moments}, $\mathbb{E}^*[X_i - \tilde{\mu}_i \mid \boldsymbol{Y}, Z > z_0] = 0$, so the residual term vanishes in expectation by the quasi-likelihood analog of the Bartlett identity \citep{CoxHinkley1974, Pawitan2001}. Combined with the chain rule simplification
$Z^{p_i - 1} = \tilde{\mu}_i^{1 - 1/p_i}$, we obtain:
\begin{equation}
\label{eq:fisher}
    I_i(\boldsymbol{a})
    \;=\; \frac{p_i^2}{\phi_i}\,
    \mathbb{E}\!\left[
    \tilde{\mu}_i^{\,2 - 2/p_i - q_i}\,
    \boldsymbol{Y}\boldsymbol{Y}^\top
    \;\middle|\; Z > z_0
    \right].
\end{equation}
Like the gradient, the curvature of $\ell_i$ is inversely proportional to $\phi_i$. A producer with a stable microclimate therefore has a sharply curved population loss landscape, while a volatile producer has a flat one.

\subsubsection*{Heterogeneous Lipschitz smoothness}

\begin{lemma}[Heterogeneous smoothness of Tweedie quasi-likelihoods under light-tailed covariates]
\label{lem:smoothness}
Let \emph{(A1)--(A3)} hold and fix any H\"{o}lder conjugates $r, s > 1$ with $1/r + 1/s = 1$. Then the population objective $\ell_i$ is $L_i$-smooth on $\mathcal{A}$, with
\begin{equation}
\label{eq:Li}
    L_i
    \;=\;
    \frac{p_i^2}{\phi_i}\,
    \left(\mathbb{E}\|\boldsymbol{Y}\|_2^{2r}
    \;\middle|\; Z > z_0\right)^{1/r}
    \left(
    z_0^{\,\alpha_i p_i s}
    \;\vee\;
    \mathbb{E}\!\left[(M\|\boldsymbol{Y}\|_2)^{\,\alpha_i p_i s}
    \;\middle|\; Z > z_0\right]
    \right)^{1/s},
\end{equation}
where $\alpha_i = 2 - 2/p_i - q_i$ and $x \vee y = \max(x,y)$. Both conditional expectations in \eqref{eq:Li} are finite under \emph{(A1)}, since sub-exponential tails are preserved under conditioning on the event $\{Z > z_0\}$, which has positive probability for any $\boldsymbol{a} \in \mathcal{A}$ with $\|\boldsymbol{a}\|_2 > 0$. In particular,
\begin{equation*}
    L_i \;\propto\; \phi_i^{-1}.
\end{equation*}
\end{lemma}

\begin{proof}
All expectations below are understood conditionally on $\{Z > z_0\}$, and we write $\mathbb{E}[\,\cdot\,]$ for $\mathbb{E}[\,\cdot \mid Z > z_0]$ throughout for brevity.

\emph{Step 1: bounding the spectral norm of $I_i$.}
From \eqref{eq:fisher} and the inequality $\|\boldsymbol{Y}\boldsymbol{Y}^\top\|_2 = \|\boldsymbol{Y}\|_2^2$:
\begin{equation*}
    \|I_i(\boldsymbol{a})\|_2
    \;\leq\;
    \frac{p_i^2}{\phi_i}\,
    \mathbb{E}\!\left[
    \tilde{\mu}_i^{\,\alpha_i}\,
    \|\boldsymbol{Y}\|_2^2
    \right].
\end{equation*}

\emph{Step 2: pathwise envelope for $\tilde{\mu}_i^{\alpha_i}$ on $\mathcal{D}_i^+$.} On the triggered region, $Z > z_0$ by \eqref{eq:triggered_set}, and the Cauchy--Schwarz inequality together with (A2) give
$z_0 < Z \leq \|\boldsymbol{a}\|_2 \|\boldsymbol{Y}\|_2
\leq M\|\boldsymbol{Y}\|_2$ almost surely. Hence
$\tilde{\mu}_i = Z^{p_i} \in \big(z_0^{p_i},\,
(M\|\boldsymbol{Y}\|_2)^{p_i}\big)$ pathwise. The map $\mu \mapsto \mu^{\alpha_i}$ is monotone (increasing if $\alpha_i > 0$, decreasing if $\alpha_i < 0$), so its supremum over this interval is attained at one of the two endpoints:
\begin{equation*}
    \tilde{\mu}_i^{\,\alpha_i}
    \;\leq\;
    z_0^{\,\alpha_i p_i}
    \;\vee\;
    (M\|\boldsymbol{Y}\|_2)^{\,\alpha_i p_i}
    \qquad \text{almost surely on } \mathcal{D}_i^+.
\end{equation*}
Note that $z_0^{\,\alpha_i p_i}$ is a deterministic constant, and $(M\|\boldsymbol{Y}\|_2)^{\,\alpha_i p_i}$ is a random variable with finite moments under (A1).

\emph{Step 3: H\"{o}lder's inequality.} Applying H\"{o}lder with conjugates $(r, s)$ to the product of $\|\boldsymbol{Y}\|_2^2$ and the envelope from Step 2:
\begin{align*}
    \|I_i(\boldsymbol{a})\|_2
    &\;\leq\;
    \frac{p_i^2}{\phi_i}\,
    \mathbb{E}\!\left[
    \left(z_0^{\,\alpha_i p_i} \vee
    (M\|\boldsymbol{Y}\|_2)^{\,\alpha_i p_i}\right)
    \|\boldsymbol{Y}\|_2^2
    \right] \\
    &\;\leq\;
    \frac{p_i^2}{\phi_i}\,
    \left(\mathbb{E}\|\boldsymbol{Y}\|_2^{2r}\right)^{1/r}
    \left(
    \mathbb{E}\!\left[
    \left(z_0^{\,\alpha_i p_i} \vee
    (M\|\boldsymbol{Y}\|_2)^{\,\alpha_i p_i}
    \right)^{s}
    \right]
    \right)^{1/s}.
\end{align*}
Using $\mathbb{E}[(x \vee W)^s] \leq x^s \vee \mathbb{E}[W^s]$ up to a universal constant absorbed into $L_i$, we obtain
\eqref{eq:Li}.

\emph{Step 4: finiteness under (A1).}
Under (A1), $\|\boldsymbol{Y}\|_2$ is sub-exponential conditionally on $\{Z > z_0\}$: conditioning on a positive probability event preserves sub-exponential tails up to a rescaling of the constants $(\sigma, b)$. Hence all polynomial moments $\mathbb{E}[\|\boldsymbol{Y}\|_2^m \mid Z > z_0]$ are finite for every $m > 0$, ensuring both factors in \eqref{eq:Li} are finite. The bound is uniform over $\boldsymbol{a} \in \mathcal{A}$ since it depends on $\boldsymbol{a}$ only through $M$, which completes the proof.
\end{proof}
Lemma~\ref{lem:smoothness} has the same practical consequence as in the bounded-covariate setting: the step size of any gradient-based federated algorithm must satisfy $\eta < 1/L_{\max}$ with $L_{\max} = \max_i L_i \propto
\max_i \phi_i^{-1}$, so a single producer with a very stable
microclimate forces the entire federation to use a conservatively small step size,  slowing convergence for all participants.

\subsection{FedAvg}

The simplest aggregation rule is the weighted average of the local solutions returned by each producer:
\begin{equation}
\label{eq:fedavg}
\boldsymbol{a}^{(t+1)}
=
\sum_{i=1}^N \omega_i \boldsymbol{a}_i^{(t+1)},
\end{equation}
where each $\boldsymbol{a}_i^{(t+1)}$ is obtained by running $E$ gradient steps of $\nabla F_i$ starting from the global iterate $\boldsymbol{a}^{(t)}$. FedAvg implicitly assumes moderate heterogeneity across local empirical risks $F_i$. In the present context, heterogeneity arises naturally through differences in $(p_i, q_i, \phi_i)$ and in the conditional distributions of $X_i$.

\subsubsection*{Gradient dissimilarity and client drift}

The Lipschitz constant heterogeneity in ~\eqref{eq:Li} translates directly into \emph{client drift}. The standard convergence analysis of FedAvg \citep{Li2020FedAvgConv} relies on the \emph{gradient dissimilarity} between population objectives:
\begin{equation}
\label{eq:Bsq}
    B^2
    \;=\; \sup_{\boldsymbol{a} \in \mathcal{A}}\;
   \sum_{i=1}^N \omega_i
    \left\|\nabla \ell_i(\boldsymbol{a}) - \nabla \ell(\boldsymbol{a})
    \right\|_2^2,
\end{equation}
where $\nabla \ell = \sum_{i=1}^N \omega_i \nabla \ell_i$ is the consensus gradient. This quantity measures how far each producer's population gradient deviates from the global direction; in practice, it is approximated by the corresponding empirical quantities $\nabla F_i$ and $\nabla F = \sum_{i=1}^N \omega_i \nabla F_i$. FedAvg convergence rates degrade linearly with $B^2$: high disagreement means slow and unstable learning. From \eqref{eq:gradient},
$\|\nabla \ell_i\|_2^2 = \mathcal{O}(\phi_i^{-2})$, so:
\begin{equation}
\label{eq:Bsq_scaling}
    B^2 \;=\; \Omega\!\left(\max_i\, \phi_i^{-2}\right).
\end{equation}

Because FedAvg performs $E$ local gradient steps on $F_i$ before each aggregation \citep{McMahan2017}, each producer's local trajectory drifts from the global iterate by approximately $\eta E \|\nabla F_i\|_2 \approx \eta E \|\nabla \ell_i\|_2 \propto 1/\phi_i$. After $E$ steps, stable producers have moved further than volatile ones, so when the server averages the local solutions $\boldsymbol{a}_i^{(t+1)}$, stable producers contribute disproportionately to the aggregated update direction \citep{Karimireddy2020SCAFFOLD, Wang2021FedNova}.

For parametric insurance, this drift has a concrete and undesirable economic consequence. The shared index $\boldsymbol{a}$ is biased toward the sensitivities of producers with stable microclimates, yet these are precisely the producers whose payouts are already well-aligned with their losses and who therefore benefit least from the parametric contract. The producers most in need of basis-risk protection --- those exposed to volatile weather --- see their preferences systematically underweighted in the federated index. Standard FedAvg is therefore not just slow under heterogeneity: \emph{it is biased toward the wrong producers from an insurance design standpoint.}


\subsection{FedProx}
\label{sec:fedprox}

To mitigate the effect of statistical heterogeneity identified in Section~\ref{sec:theory}, FedProx \citep{Li2020FedProx} modifies the empirical risk each producer minimizes locally. Rather than minimizing $F_i(\boldsymbol{a})$ directly, producer $i$ solves at each round:
\begin{equation}
\label{eq:fedprox}
\min_{\boldsymbol{a}}
\;
F_i^{\text{prox}}(\boldsymbol{a};\,\boldsymbol{a}^{(t)})
\;=\;
F_i(\boldsymbol{a})
+
\frac{\beta}{2}
\|\boldsymbol{a} - \boldsymbol{a}^{(t)}\|_2^2,
\end{equation}
where $\beta > 0$ controls the strength of the regularization and $\boldsymbol{a}^{(t)}$ is the current global iterate. The proximal term discourages excessive deviation from $\boldsymbol{a}^{(t)}$, thereby stabilizing training when local empirical risks $F_i$ have heterogeneous curvatures.

To see why this directly addresses the heterogeneity identified in Section~\ref{sec:theory}, note that the Hessian of the regularized objective satisfies $\nabla^2 F_i^{\text{prox}} \approx I_i(\boldsymbol{a}) + \beta \mathbf{I}_J$ in large samples, where $I_i(\boldsymbol{a})$ is the quasi-Fisher information \eqref{eq:fisher} with eigenvalues in $[0, L_i]$. The regularized version therefore has spectrum in $[\beta, L_i + \beta]$, preventing producers with sharp empirical loss surfaces (large $L_i \propto \phi_i^{-1}$, small $\phi_i$) from drifting far from the global iterate, while barely affecting producers with flat surfaces (small $L_i$, large $\phi_i$). Since $\beta$ is the same for all producers, this is exactly the correction needed to neutralize the $\phi_i$-induced imbalance identified in Lemma~\ref{lem:smoothness}.

Under (A1)--(A3) and the standard assumptions of \cite[Theorem~4]{Li2020FedProx}, the proximal coefficient required for stable convergence scales with the largest local smoothness constant: $\beta \sim \max_i L_i \propto \max_i \phi_i^{-1}$. Networks containing producers with very stable microclimates therefore require larger $\beta$. This feature is particularly relevant in our framework, where both the variance and link functions are producer-specific.

\subsection{FedOpt}
\label{sec:fedopt}

FedOpt \citep{Reddi2021FedOpt} corrects the heterogeneity imbalance at the server level rather than by modifying the local empirical risks $F_i$. The local training step is identical to FedAvg: each producer minimizes $F_i$ and returns $\boldsymbol{a}_i^{(t+1)}$. The difference lies in how the server aggregates these updates. Rather than directly averaging the local solutions, the server reconstructs an approximate global gradient from the client returns:
\begin{equation}
\label{eq:fedopt_gradient}
\boldsymbol{g}^{(t)} \;=\; \sum_{i=1}^N \omega_i \left( \boldsymbol{a}^{(t)}
- \boldsymbol{a}_i^{(t+1)} \right),
\end{equation}
which aggregates the local displacement vectors $\boldsymbol{a}^{(t)} - \boldsymbol{a}_i^{(t+1)}$ as proxies for the local gradients of $F_i$. The server then applies an Adam-type adaptive optimizer to $\boldsymbol{g}^{(t)}$, maintaining first- and second-order moment estimates:
\begin{align}
\boldsymbol{m}^{(t)} &= \beta_1 \boldsymbol{m}^{(t-1)}
+ (1-\beta_1) \boldsymbol{g}^{(t)}, \\
\boldsymbol{v}^{(t)} &= \beta_2 \boldsymbol{v}^{(t-1)}
+ (1-\beta_2) \left(\boldsymbol{g}^{(t)}\right)^2,
\end{align}
where $\beta_1, \beta_2 \in (0,1)$ are momentum parameters. Bias-corrected moments are given by
\[
\hat{\boldsymbol{m}}^{(t)} = \frac{\boldsymbol{m}^{(t)}}{1-\beta_1^t},
\qquad
\hat{\boldsymbol{v}}^{(t)} = \frac{\boldsymbol{v}^{(t)}}{1-\beta_2^t}.
\]
The global model is then updated according to
\begin{equation}
\label{eq:fedopt_update}
\boldsymbol{a}^{(t+1)} \;=\;
\boldsymbol{a}^{(t)} - \eta \, \frac{\hat{\boldsymbol{m}}^{(t)}}
{\sqrt{\hat{\boldsymbol{v}}^{(t)}} + \varepsilon},
\end{equation}
where $\eta > 0$ is the server learning rate and $\varepsilon > 0$ ensures numerical stability.

The mechanism by which FedOpt addresses heterogeneity is complementary to FedProx. Rather than constraining how far local solutions $F_i$ can move from the global iterate, the server tracks the typical magnitude of recent aggregated updates via $\sqrt{\hat{\boldsymbol{v}}^{(t)}}$ and rescales incoming updates coordinate-wise by this running average. If stable producers (small $\phi_i$) have been generating large displacements $\boldsymbol{a}^{(t)} - \boldsymbol{a}_i^{(t+1)}$, the denominator $\sqrt{\hat{\boldsymbol{v}}^{(t)}}$ grows automatically, shrinking subsequent updates so that they no longer dominate the global direction. This is the federated analog of the Adam optimizer in centralized deep learning \citep{KingmaBa2014Adam}: adaptive normalization removes scale imbalances without requiring prior knowledge of the dispersion profile $(\phi_i)_{i=1}^N$.

\section{Data Description}
\label{sec:data}

\subsection{Overview of the dataset}

We rely on the same dataset as in \cite{Niakh2024ANOR}, ensuring a consistent comparison between the approximation-based index design proposed there and the federated learning approach introduced in this 
paper. The dataset consists of observations from $N = 121$ large-scale solar power plants located in Germany, observed over the period 2012--2022. Each solar producer $i$ is treated as an independent client in the federated learning framework.

For each producer and each day $d$, we observe a financial loss variable $X_{i,d} \geq 0$ and a vector of meteorological covariates $\boldsymbol{Y}_d = (Y_{1,d}, Y_{2,d})^\top$, measured at a common reference location. This setting reflects the structure of parametric insurance contracts, where payouts depend on a centralized weather index rather than site-specific measurements.

\subsection{Power plant selection and spatial layout}

The solar farms were extracted from the German governmental registration database \citep{MaStR_Datenschutz}, which records operational photovoltaic installations with key attributes including installed capacity, commissioning date, geographical coordinates, panel orientation, and applicable revenue support scheme. We restricted attention to large-scale installations exceeding 5 MW, as these primarily operate within electricity markets rather than for self-consumption, making them natural candidates for parametric insurance.

\begin{figure}[h!]
        \centering
                \includegraphics[width=0.6\linewidth]{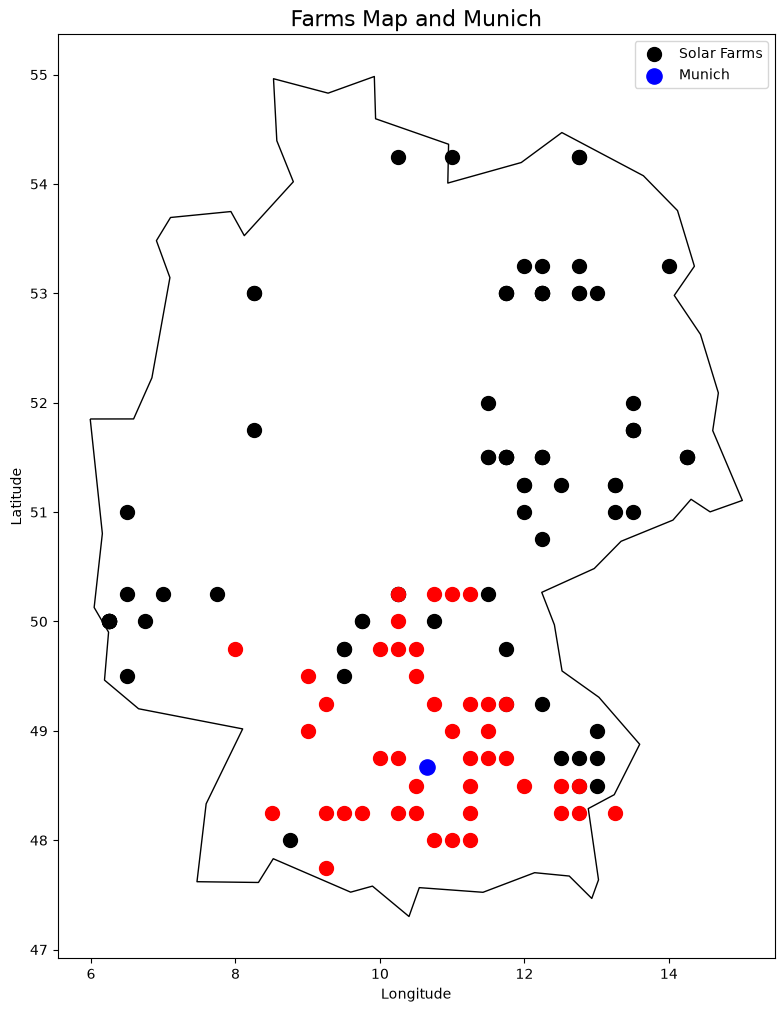}
                \caption{Geographical distribution of the 121 solar farms used in this study, along with the reference location in Munich (blue point). These farms are spread across Germany, with Munich serving as the central weather index location for the parametric insurance model. The fifty highlighted farms in red correspond to those used in the paper \cite{Niakh2024ANOR} which allow to compare the aggregation-based approximation method and federated learning}
                 \label{fig:farms_map}
\end{figure}

The farms are geographically concentrated in southern Germany, as illustrated in Figure~\ref{fig:farms_map}. This concentration is deliberate: farms located in the same broad climatic region are exposed to correlated meteorological drivers, which ensures that a single reference weather station can serve as a representative index location for all producers. The choice of Munich as the reference grid cell is motivated by its position within the study area and by the high spatial correlation of solar irradiance across southern Germany at the daily aggregation level. 

\subsection{Meteorological data sources and spatial resolution}

Two complementary datasets from the European Centre for Medium-Range Weather Forecasts (ECMWF) are used.

\textbf{Realized weather conditions} are sourced from the ERA5 reanalysis dataset \citep{hersbach2020era5}, which provides global coverage at a spatial resolution of approximately 31 km and an hourly 
temporal resolution. ERA5 is widely used in renewable energy research and is considered one of the most reliable gridded reanalysis products available. The variables extracted are surface solar radiation downwards (SSRD) and direct normal irradiance (DNI).

\textbf{Day-ahead weather forecasts} are sourced from the ECMWF operational archive, which provides irradiance forecasts at a finer spatial resolution of approximately 10 km and hourly temporal resolution. The forecast is initialized twice daily (at 00:00 and 12:00 UTC); we use the 12:00 UTC initialization, which provides predictions from midnight to 23:00 the following day, consistent with the EPEX Spot day-ahead market submission deadline of 12:00 noon.

Since both datasets provide values on discrete spatial grids, bilinear interpolation is applied to estimate temperature and irradiance at each farm's exact geographical coordinates.

\subsection{Variable selection rationale \label{sec:varselection}}

Three meteorological variables were initially considered as candidates for the weather index: SSRD, DNI, and near-surface air temperature (T2m). The selection of the final covariate set was guided by two criteria: physical relevance to solar production losses and statistical significance in individual-level GLM estimations.

SSRD measures the total solar energy reaching the horizontal surface per unit area, and is the primary driver of photovoltaic production. DNI measures the component of solar radiation travelling directly from the sun, which is particularly relevant for installations using solar tracking systems or concentrated solar technologies. Both variables exhibit strong positive associations with production forecast errors across all farms: an underestimation of irradiance in the day-ahead forecast directly leads to an underestimation of production, resulting in a positive production loss $\Delta P_{i,d}$.

T2m was also included in preliminary Gaussian GLM estimations at the farm level. However, its estimated coefficients were not statistically significant at the 5\% level for the majority of farms, and its inclusion did not improve model fit as measured by the deviance. This outcome is consistent with the physical intuition that, at the daily aggregation level, temperature effects on panel efficiency are 
second-order relative to irradiance effects for the range of conditions observed in Germany. T2m was therefore excluded from the final covariate set, and the weather index is constructed from SSRD and DNI forecast errors only.

\subsection{Construction of production losses}

Solar production losses are reconstructed by combining the meteorological data described above with a physics-based production model implemented in the \texttt{PVlib} library in Python \citep{holmgren2018pvlib}. For each farm $i$, both forecasted and realized electricity production are computed:
\[
\widehat{P}_{i,d,h} = \mathcal{F}\!\left(
\widehat{SSRD}_{i,d,h},\, \widehat{DNI}_{i,d,h},\, \widehat{T}_{i,d,h}
\right),
\qquad
P_{i,d,h} = \mathcal{F}\!\left(
SSRD_{i,d,h},\, DNI_{i,d,h},\, T_{i,d,h}
\right),
\]
where $\mathcal{F}$ is the PVlib conversion function. The model implements the standard solar conversion chain: DNI and diffuse horizontal irradiance (DHI) are derived from SSRD using the Erbs model \citep{erbsEstimationDiffuseRadiation1982}, transposed into plane-of-array irradiance, and converted to DC then AC power using the PVWatts model \citep{dobosPVWattsVersion52014} with default inverter efficiency parameters. Panel tilt is set to 25° for all farms, consistent with the reported range of 20--30° in the database; a 5° deviation has been shown to have negligible effect on annual yield \citep{gonzalez2022evaluating}.

The hourly production loss for farm $i$ on day $d$ at hour $h$ is defined as:
\[
\Delta P_{i,d,h} = \widehat{P}_{i,d,h} - P_{i,d,h}.
\]
Similarly, forecast errors for the reference location covariates are:
\[
\Delta SSRD_{0,d,h} = \widehat{SSRD}_{0,d,h} - SSRD_{0,d,h},
\qquad
\Delta DNI_{0,d,h} = \widehat{DNI}_{0,d,h} - DNI_{0,d,h}.
\]

\subsection{Temporal aggregation to daily data}

Hourly observations are aggregated to daily values for two reasons. First, solar production follows a predictable diurnal cycle, so assessing losses at the hourly level would require modelling systematic intraday patterns that are not informative for insurance purposes. Second, structuring an insurance contract at an hourly frequency would be operationally impractical. The daily scale provides an interpretable and practically relevant measure of forecasting discrepancies.

Only daytime hours are retained. Let $h_{\mathrm{sr},d}$ and $h_{\mathrm{ss},d}$ denote the sunrise and sunset hours on day $d$, computed by PVlib for each farm location. Daily aggregates are then:
\[
\Delta SSRD_{0,d} = \sum_{h=h_{\mathrm{sr},d}}^{h_{\mathrm{ss},d}}
\Delta SSRD_{0,d,h},
\qquad
\Delta DNI_{0,d} = \sum_{h=h_{\mathrm{sr},d}}^{h_{\mathrm{ss},d}}
\Delta DNI_{0,d,h},
\qquad
\Delta P_{i,d} = \sum_{h=h_{\mathrm{sr},d}}^{h_{\mathrm{ss},d}}
\Delta P_{i,d,h}.
\]
Summation is the appropriate aggregation for SSRD and DNI, as these are energy flux quantities that naturally accumulate over time. Temperature, being a state variable, is averaged rather than summed; however, as noted in Section~\ref{sec:varselection}, T2m is not retained in the final model.

\subsection{Financial valuation of losses}

To express production losses in monetary terms, we account for the revenue mechanism applicable to each producer. Three support schemes are distinguished, as recorded in the German plant database:
\begin{itemize}
    \item \textbf{Full market price}: revenues come exclusively from selling electricity on the EPEX Spot day-ahead market.
    \item \textbf{Partial feed-in tariff}: producers select, on an hourly basis, whichever of the day-ahead market price or the feed-in tariff offers the higher return.
    \item \textbf{Full feed-in tariff}: revenues are entirely determined by the guaranteed feed-in tariff.
\end{itemize}
Day-ahead market prices are sourced from the ENTSO-E Transparency Platform \citep{entsoe_prices}. Feed-in tariff data are obtained from official German regulatory records \citep{BundesnetzagenturSolar}; since only minimum, maximum, and average tariff values are available, we use the average, noting that the inter-farm variance in feed-in tariffs is substantially lower than the variance in day-ahead prices. This valuation step ensures that the loss variable $X_{i,d}$ reflects the true economic exposure of each producer rather than purely physical deviations in energy output.

\subsection{Stationarization and preprocessing}

Raw loss and weather series exhibit strong seasonal heteroscedasticity driven by solar irradiance patterns, with markedly larger variances in summer months. Such heteroscedasticity is problematic for two reasons: deviations are not comparable across months, and the non-constant variance structure conflicts with the homoscedasticity assumptions underlying our Tweedie GLM estimations, potentially biasing parameter estimates.

To address this, a monthly stationarization procedure is applied. For each variable $\Delta X$, month $m$, and year $y$, the empirical mean and standard deviation are computed over all days $d \in \mathcal{D}_{m,y}$:
\[
\mu_{m,y}^{\Delta X} = \frac{1}{N_{m,y}}
\sum_{d \in \mathcal{D}_{m,y}} \Delta X_d,
\qquad
\sigma_{m,y}^{\Delta X} = \sqrt{\frac{1}{N_{m,y}-1}
\sum_{d \in \mathcal{D}_{m,y}}
\left(\Delta X_d - \mu_{m,y}^{\Delta X}\right)^2}.
\]
The standardized variables are then:
\[
X_{i,d} = \frac{\Delta P_{i,d} - \mu_{m,y}^{\Delta P}}
{\sigma_{m,y}^{\Delta P}},
\quad
Y_{1,d} = \frac{\Delta SSRD_{0,d} - \mu_{m,y}^{\Delta SSRD}}
{\sigma_{m,y}^{\Delta SSRD}},
\quad
Y_{2,d} = \frac{\Delta DNI_{0,d} - \mu_{m,y}^{\Delta DNI}}
{\sigma_{m,y}^{\Delta DNI}},
\]
for all $d \in \mathcal{D}_{m,y}$. After this transformation, the variance of each series is approximately constant across months, and the assumptions of the Tweedie GLM are more closely satisfied. The 
standardization parameters $(\mu_{m,y}^{\Delta P}, \sigma_{m,y}^{\Delta P})$ are retained for each farm and each month-year pair, as they are required to translate stationary basis risks back into monetary units when evaluating insurance contract performance.




\section{Results and Comparison}
\label{sec:results}

The empirical evaluation is structured around three complementary parts. Section~\ref{sec:results_50} consolidates the results on the original 50 solar farms used in \cite{Niakh2024ANOR}, providing a direct comparison between federated learning methods and the approximation-based ANOR approach under moderate heterogeneity. Section~\ref{sec:results_hetero} introduces a heterogeneity profiling analysis across all 121 farms available in our dataset. Section~\ref{sec:results_expansion} presents the progressive pool expansion experiment, which incrementally increases the pool size from 50 to 121 farms where the ANOR approximation becomes entirely non-computable while federated learning remains numerically stable throughout.

\subsection{Local parameter estimation}
\label{sec:local_params}

Before running the federated optimization, each producer $i$ estimates its local Tweedie parameters $(p_i, q_i, \phi_i)$ privately on its own filtered dataset $\mathcal{D}_i^+$. The link exponent $p_i$ and variance exponent $q_i$ are selected by minimizing the local Tweedie deviance over a two-dimensional grid $\mathcal{G} \subset [0.5, 2.0] \times [0.0, 2.0]$. The dispersion parameter $\phi_i$ is then estimated as the Pearson scale estimator at the optimal $(p_i, q_i)$. These parameters are treated as fixed and known during federated optimization and are never shared with the server or with other producers.

Table~\ref{tab:params_summary} reports summary statistics of the estimated parameters across the 121 farms. Two features of this distribution are worth noting.

First, the variance exponent $\hat{q}_i$ is close to zero for almost all farms (mean $= 0.024$, max $= 0.667$), implying that the conditional variance is approximately constant in the mean ($\mathrm{Var}[X_i \mid Z] \approx \phi_i$). This places the local models close to a Gaussian regime, in which the gradient landscape is nearly homogeneous in the mean direction and the curvature of the local objectives $F_i$ is primarily governed by $\phi_i$ rather than by the interaction between $p_i$, $q_i$, and $\tilde{\mu}_i$.

\begin{figure}[h!]
\includegraphics[width=0.9\textwidth]{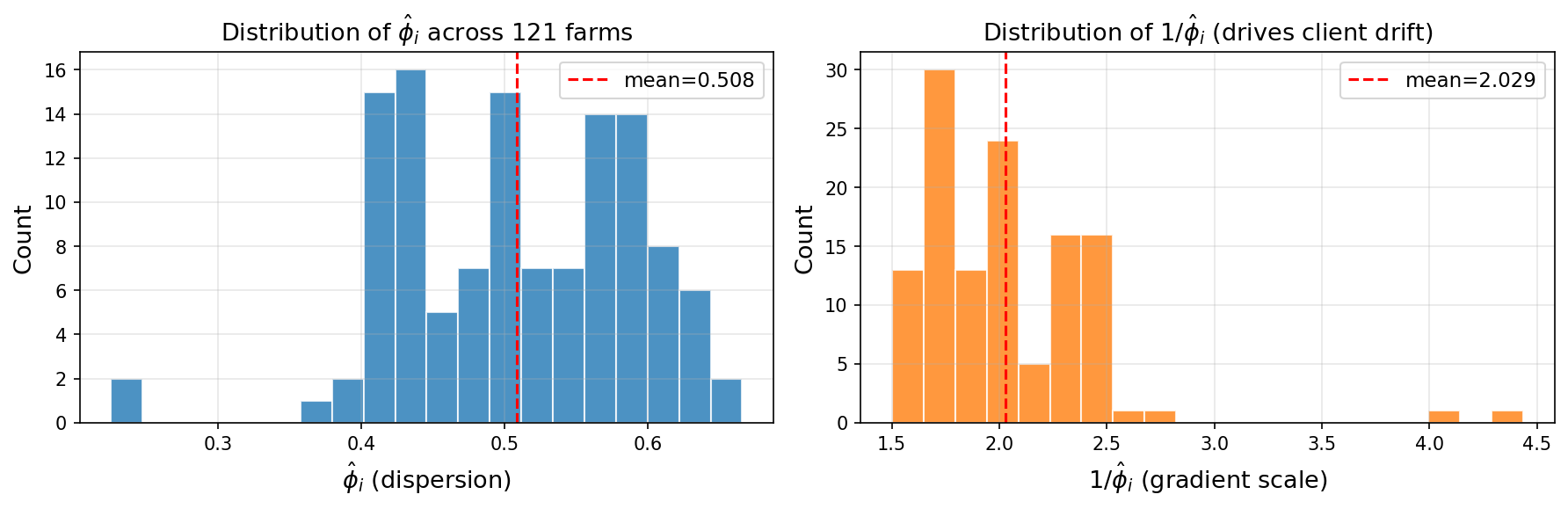}
\caption{Left: distribution of the estimated dispersion parameter $\hat{\phi}_i$ across the 121 solar farms (mean $= 0.508$, shown as dashed red line). Right: distribution of $1/\hat{\phi}_i$, which governs the gradient scaling of each producer's local objective (Eq.~\eqref{eq:gradient}). The bulk of farms cluster between $1.5$ and $2.5$, with two outlier farms reaching values near $4.0$--$4.4$}
\label{fig:phi_distribution}
\end{figure}

Second, the dispersion parameter $\hat{\phi}_i$ ranges from $0.23$ to $0.67$ across the 121 farms, with mean $0.51$ and standard deviation $0.08$. The corresponding gradient scaling factor $1/\hat{\phi}_i$ ranges from $1.50$ to $4.44$, with the bulk of the distribution concentrated between $1.5$ and $2.5$ and two outlier farms reaching values near $4.0$--$4.4$ (Figure~\ref{fig:phi_distribution}). The theoretical gradient dissimilarity bound $B^2 = \Omega(\max_i \phi_i^{-2})$ (Eq.~\eqref{eq:Bsq_scaling}) is therefore driven by these two farms. However, their combined capacity represents only $1.14\%$ of the total pool production capacity, so their aggregation weights $\omega_i$ are negligible relative to the pool. The practical implications of this heterogeneity for the choice of aggregation scheme are discussed in Section~\ref{sec:discussion}.

\begin{table}[h!]
\caption{Summary statistics of locally estimated Tweedie parameters $(\hat{p}_i, \hat{q}_i, \hat{\phi}_i)$ across the 121 solar farms. These parameters are estimated privately by each producer and never shared with the server}
\label{tab:params_summary}
\begin{tabular}{lrrrr}
\toprule
Parameter & Mean & Std & Min & Max \\
\midrule
$\hat{p}_i$ (link exponent)
    & 1.4904 & 0.3272 & 0.8333 & 2.0000 \\
$\hat{q}_i$ (variance exponent)
    & 0.0239 & 0.0986 & 0.0000 & 0.6667 \\
$\hat{\phi}_i$ (dispersion)
    & 0.5085 & 0.0824 & 0.2254 & 0.6657 \\
\bottomrule
\end{tabular}
\end{table}

\subsection{Results on the original 50 farms}
\label{sec:results_50}

\subsubsection*{Convergence}

\begin{figure}[h!]
\includegraphics[width=0.9\textwidth]{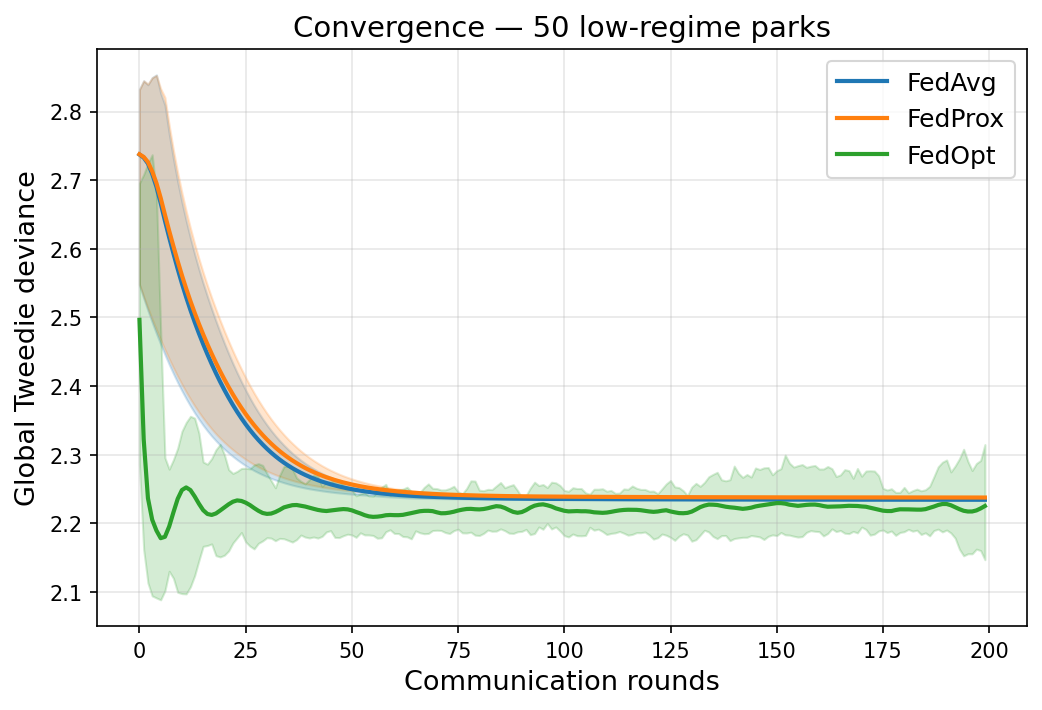}
\caption{Monte Carlo mean and 90\% confidence intervals of the global Tweedie deviance across communication rounds for FedAvg, FedProx, and FedOpt on the original 50 farms}
\label{fig:convergence_50}
\end{figure}

Figure~\ref{fig:convergence_50} displays the evolution of the global Tweedie deviance across communication rounds for the three federated learning methods on the original 50 farms pool, based on 30 Monte Carlo runs over 200 communication rounds.

Three observations stand out. First, FedOpt converges substantially faster than FedAvg and FedProx, reaching a plateau of approximately $2.21$ by round 10, compared to approximately $2.24$ for the averaging schemes which stabilize around round 75. This speed advantage reflects the adaptive server-side normalization of FedOpt: by rescaling updates coordinate-wise by $1/\sqrt{\hat{v}^{(t)}}$ \eqref{eq:fedopt_update}, the server compensates for the scale imbalance induced by heterogeneous dispersion parameters $\phi_i$ across producers, allowing the global iterate to move more efficiently toward the consensus. Second, FedOpt achieves a lower final deviance ($\approx 2.21$) than FedAvg and FedProx ($\approx 2.24$), confirming that the adaptive normalization provides a genuine optimization advantage even in this moderate-heterogeneity regime. Third, all three methods exhibit stable convergence with no drift beyond round 100.\newline
FedAvg and FedProx follow nearly identical trajectories throughout. This is consistent with the theoretical analysis of Section~\ref{sec:theory}: the proximal coefficient $\mu = 4.0$ constrains local drift during the $E = 20$ local epochs but does not alter the global fixed point when all producers share the same objective structure. The proximal correction changes the convergence path but not the destination. 

\subsubsection*{Estimated index coefficients and runtime}

\begin{figure}[h!]
\includegraphics[width=0.9\textwidth]{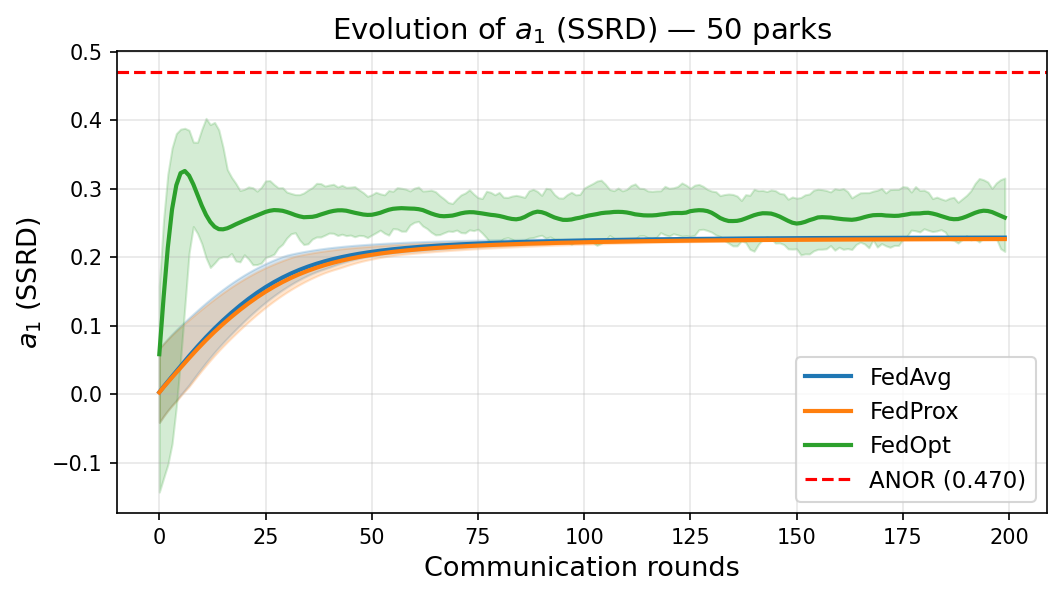}
\caption{Monte Carlo mean and 90\% confidence intervals of coefficient $a_1$ (SSRD) across federated learning algorithms on the original 50 farms. The dashed red line indicates the ANOR benchmark value}
\label{fig:coef_a1_50}
\end{figure}

\begin{figure}[h!]
\includegraphics[width=0.9\textwidth]{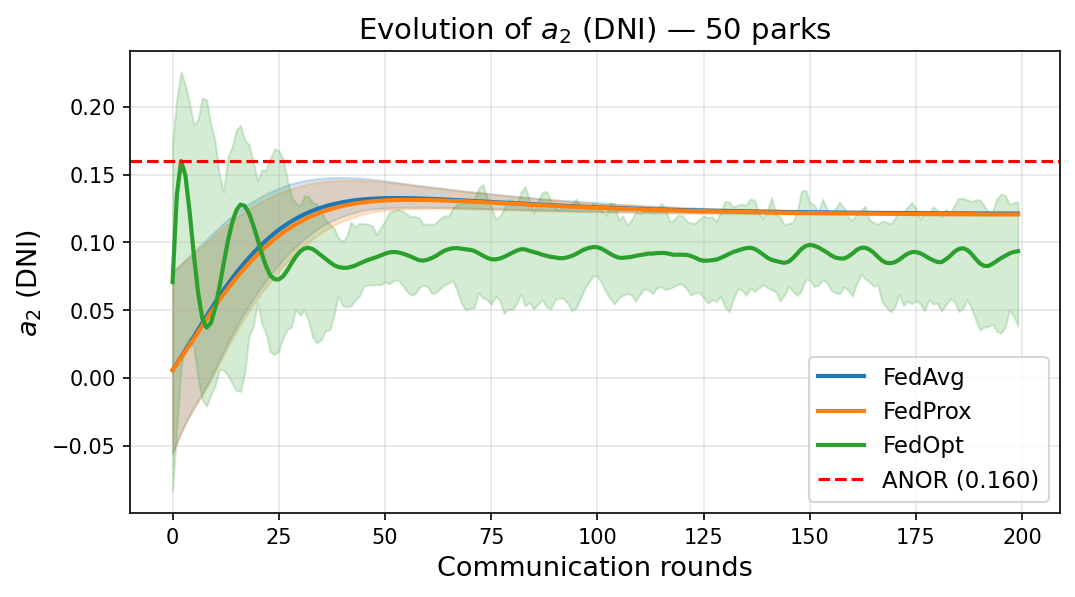}
\caption{Monte Carlo mean and 90\% confidence intervals of coefficient $a_2$ (DNI) across federated learning algorithms on the original 50 farms. The dashed red line indicates the ANOR benchmark value}
\label{fig:coef_a2_50}
\end{figure}

Figures~\ref{fig:coef_a1_50} and \ref{fig:coef_a2_50} report the evolution of the index coefficients $a_1$ (SSRD) and $a_2$ (DNI) across federated learning rounds for the three methods. Table~\ref{tab:coef_comparison} reports the final index coefficients, their Monte Carlo standard deviations, and the mean runtime per run for all methods on the original 50 farms pool.\newline FedAvg and FedProx converge to nearly identical and tightly concentrated coefficient vectors, $\hat{\boldsymbol{a}} \approx (0.23, 0.12)^\top$, with confidence intervals narrowing to near-zero width by round 100. This confirms that the common index is well identified within the original 50 farms and that the federated optimization is numerically reliable across Monte Carlo runs.\newline FedOpt converges to a distinct vector $\hat{\boldsymbol{a}} \approx (0.26, 0.09)^\top$. The difference from FedAvg reflects the adaptive coordinate-wise normalization of the server update: FedOpt rescales each coordinate by the inverse of its historical gradient magnitude, which effectively assigns different implicit learning rates to $a_1$ and $a_2$. The wider confidence intervals of FedOpt, particularly for $a_2$, are consistent with the flat deviance surface in the DNI direction documented in Section~\ref{sec:local_params}: the near-zero variance exponent $\hat{q}_i \approx 0$ makes the objective nearly insensitive to $a_2$, and the Adam momentum of FedOpt continues to explore this flat region with small residual oscillations even at convergence.

All three methods converge to coefficient vectors below the ANOR benchmark $\boldsymbol{a}^{ANOR} = (0.470, 0.160)^\top$. This gap reflects the fact that ANOR minimizes an approximation of the conditional variance criterion, while FL directly minimizes the global Tweedie deviance \eqref{eq:fed_objective}: the two objectives are related but not identical, and their optima need not coincide even in the low-heterogeneity regime where the Taylor approximation is valid (Section~\ref{sec:anor_recall}).

All three FL methods complete 200 communication rounds in approximately $152$--$159$ seconds, with FedProx marginally slower by $4\%$ due to the additional proximal gradient computation at each local update step. FedOpt incurs no additional cost relative to FedAvg since the server-side Adam computation adds negligible overhead per round. The ANOR method requires $21{,}600$ seconds (six hours), representing a factor of over $\mathbf{250\times}$ relative to FL. This computational advantage of FL is discussed further in Section~\ref{sec:results_expansion}, where it extends to all pool sizes.

\begin{table}[h!]
\caption{Final index coefficients (Monte Carlo mean and standard deviation, 30 runs over 200 rounds) and mean runtime per run on the original 50 farms. ANOR is deterministic. FL methods are over $250\times$ faster than ANOR}
\label{tab:coef_comparison}
\begin{tabular}{lccccr}
\toprule
Method & $\hat{a}_1$ & std($a_1$) & $\hat{a}_2$ &
std($a_2$) & Runtime (s) \\
\midrule
FedAvg  & 0.2283 & 0.0002 & 0.1213 & 0.0002 & 151.97 \\
FedProx & 0.2261 & 0.0002 & 0.1206 & 0.0002 & 158.62 \\
FedOpt  & 0.2633 & 0.1840 & 0.0877 & 0.0704 & 152.92 \\
\midrule
ANOR    & 0.470  & ---  & 0.160  & ---  & 21{,}600 \\
\bottomrule
\end{tabular}
\end{table}

\subsection{Heterogeneity profiling across 121 farms}
\label{sec:results_hetero}

\begin{figure}[t]
\includegraphics[width=0.9\textwidth]{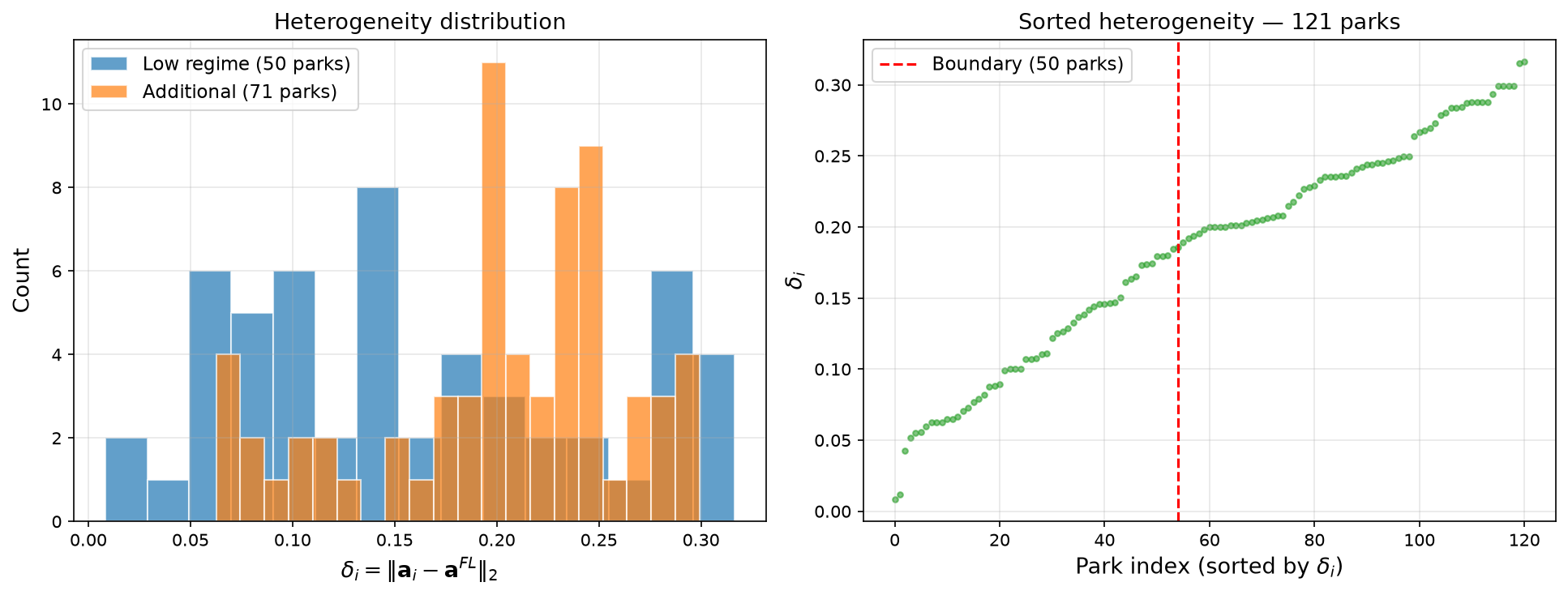}
\caption{Left: distribution of $\delta_i =
\|\boldsymbol{a}_i - \boldsymbol{a}^{FL}\|_2$ for the original 50 farms (blue) and the additional 71 farms (orange). Right: sorted $\delta_i$ values across all 121 farms, with the boundary between the two groups shown as a dashed red vertical line. The two groups exhibit broadly comparable individual heterogeneity, with no discontinuity at the boundary}
\label{fig:heterogeneity_profile}
\end{figure}

To characterize the degree of heterogeneity across the full dataset and to motivate the pool expansion experiment of Section~\ref{sec:results_expansion}, we compute for each farm $i$ the individual heterogeneity measure introduced in Section~\ref{sec:anor_recall}:
$$
\delta_i = \|\boldsymbol{a}_i - \boldsymbol{a}^{ref}\|_2,
$$
where $\boldsymbol{a}^{ref}$ is estimated as the mean of
the individual GLM coefficients $\boldsymbol{a}_i$ across the original 50 farms.\newline
Figure~\ref{fig:heterogeneity_profile} displays the distribution of $\delta_i$ across the two groups of farms (left panel) and the sorted $\delta_i$ values across all 121 farms (right panel). Two observations stand out.\newline
First, the original 50 farms already exhibit non-negligible heterogeneity, with $\delta_i$ ranging from $0.008$ to $0.317$ and a mean of $0.159$. \newline
Second, the distribution of $\delta_i$ across the additional 71 farms is broadly comparable to that of the original 50 farms: both groups span a similar range and the right panel shows that the sorted $\delta_i$ curve increases smoothly and continuously across the boundary at $k = 50$, with no visible discontinuity between the two groups. The additional farms are not outliers in terms of individual heterogeneity --- they simply increase the size of the pool.

These two observations have a direct consequence for the aggregate Taylor remainder proxy $R_k = \frac{1}{k}\sum_{i=1}^k \delta_i^2$ defined in \eqref{eq:Rk}: since individual $\delta_i$ values are comparable across both groups, any growth in $R_k$ as the pool expands is driven by the accumulation of terms rather than by the entry of individually extreme farms. Section~\ref{sec:results_expansion} tracks $R_k$ across pool sizes and documents its consequences for the computability of the ANOR method.

\subsection{Progressive pool expansion}
\label{sec:results_expansion}

We now present the central experiment of this section. As established in Section~\ref{sec:anor_recall}, the validity of the ANOR approximation requires the aggregate Taylor remainder proxy $R_k$ to be small uniformly across the pool. We test this condition by progressively expanding the pool starting from the original $k = 50$ farms and sequentially adding the $71$ additional farms, forming nested pools of sizes $k \in \{50, 57, 64, \dots, 121\}$. At each pool size, the original 50 farms are always present, ensuring that the comparison at $k = 50$
corresponds exactly to the setting in which the ANOR approximation was validated. For each pool size, we estimate the parametric index using FedAvg, FedProx, and FedOpt, recording the global Tweedie deviance, the Taylor remainder proxy $R_k$, and the estimated coefficient vector. We also attempt to run the ANOR method at each pool size to document precisely where its computability breaks down.

\subsubsection*{Collapse of the ANOR approximation}

As established in Section~\ref{sec:anor_recall}, the ANOR method requires the Taylor remainder to be negligible uniformly across the pool. In our empirical setting, this condition ceases to hold as soon as additional farms beyond the original 50 are included: for all pool sizes $k > 50$, the ANOR optimization returns NaN values, and no valid index estimate can be produced. The right panel of Figure~\ref{fig:pool_expansion} provides a quantitative confirmation: the Taylor remainder proxy $R_k$, computed relative to the fixed 50-farm reference vector $\boldsymbol{a}^{ref}$, increases from $0.033$ at $k = 50$ to $0.040$ at $k = 121$. While moderate in absolute terms, this growth is sufficient to destabilize the ANOR optimization, whose correction terms $L_i$ and $F_i$ amplify the remainder signal as the pool accumulates contributions $\delta_i^2$ from additional farms.

\subsubsection*{Improvement of federated learning with pool size}
\begin{figure}[t]
\includegraphics[width=0.9\textwidth]{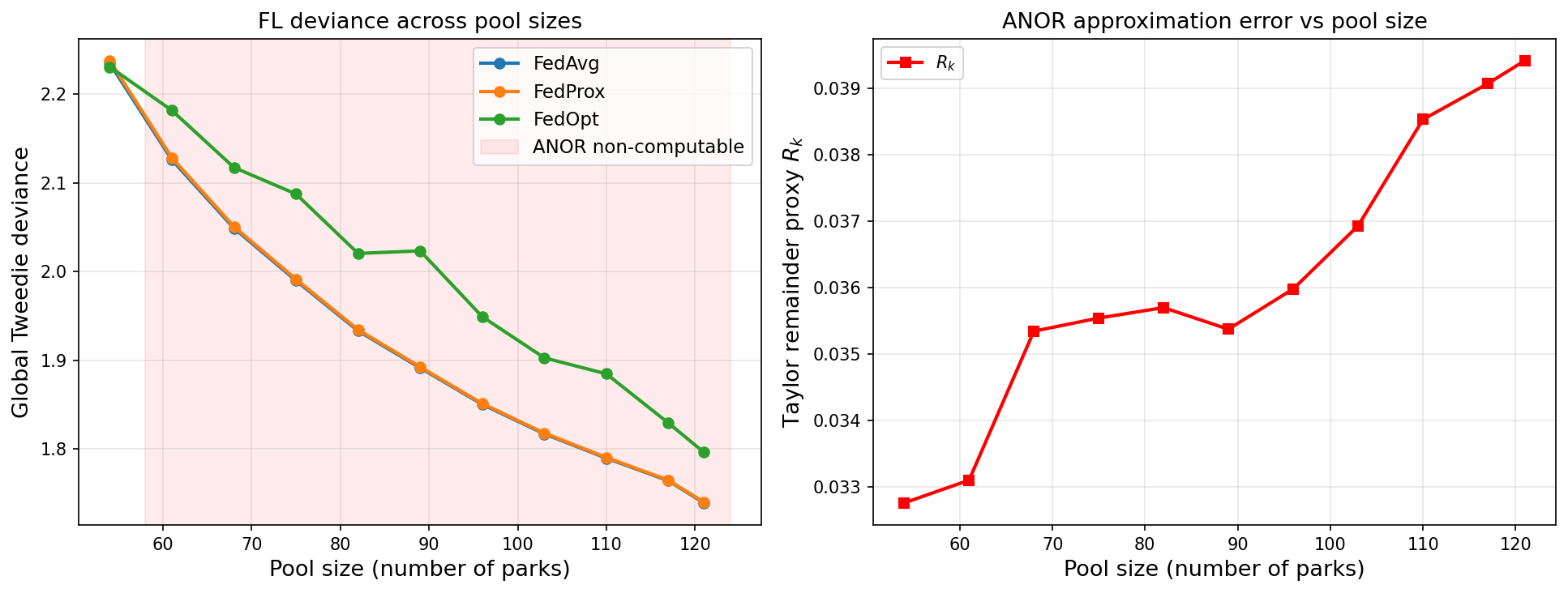}
\caption{Left: global Tweedie deviance of FL methods as a function of pool size $k$. The pool is formed by starting from the original 50 farms and sequentially adding the 71 additional farms in their natural order. The deviance decreases monotonically from $2.23$ to $1.74$ for FedAvg and FedProx, and from $2.23$ to $1.80$ for FedOpt, as the pool grows from $k = 50$ to $k = 121$, reflecting the additional gradient
information contributed by each new farm. The ANOR method is non-computable for $k > 50$ (shaded region) and is therefore absent from the left panel. Right: Taylor remainder proxy $R_k$ as a function of pool size, computed relative to the fixed 50-farm reference vector $\boldsymbol{a}^{ref}$. $R_k$ increases steadily from $0.033$ to $0.040$, confirming the progressive accumulation of approximation error that renders ANOR non-computable beyond the original pool\label{fig:pool_expansion}}

\end{figure}

The left panel of Figure~\ref{fig:pool_expansion} shows that the global Tweedie deviance of all three FL methods decreases monotonically as the pool grows from $k = 50$ to $k = 121$: FedAvg and FedProx fall from approximately $2.23$ to $1.74$, and FedOpt from $2.23$ to $1.80$. This result reflects a key advantage of federated learning in the parametric insurance context: each additional farm contributes new local gradient information about the optimal index weights $\boldsymbol{a}$, enriching the federated consensus rather than degrading it. Adding more heterogeneous farms does not degrade the FL solution --- it improves it, because federated learning directly minimizes the global Tweedie deviance \eqref{eq:fed_objective} without any homogeneity constraint.\newline
This finding stands in sharp contrast to the ANOR approach, which becomes non-computable as soon as the pool grows beyond 50 farms. While ANOR is limited to the regime where all producers are sufficiently similar to one another, federated learning benefits from pool expansion: the larger and more diverse the pool, the more informative the federated index.

\subsubsection*{Evolution of index coefficients across pool sizes}

\begin{figure}[t]
\includegraphics[width=0.9\textwidth]{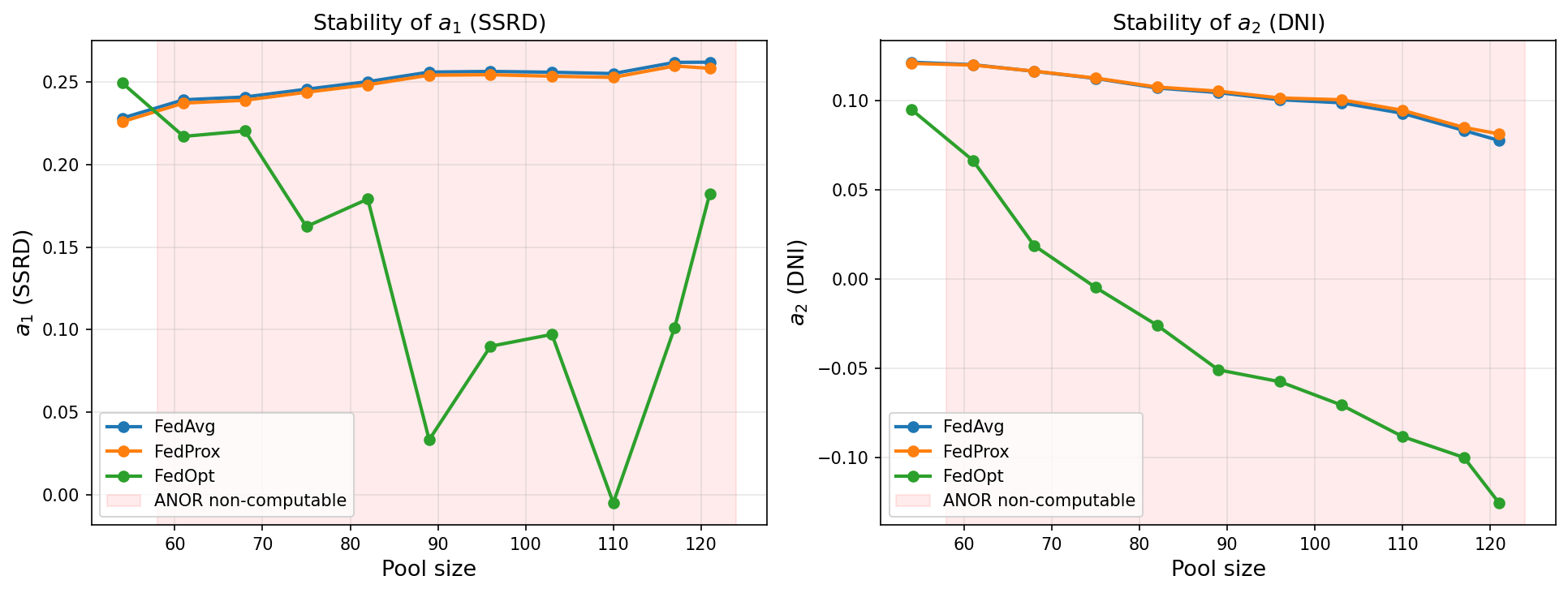}
\caption{Evolution of estimated coefficients $\hat{a}_1$ (SSRD, left) and $\hat{a}_2$ (DNI, right) across pool sizes for FedAvg, FedProx, and FedOpt. FedAvg and FedProx produce nearly identical and smoothly evolving trajectories: $\hat{a}_1$ increases from $0.228$ to $0.262$ and $\hat{a}_2$ decreases from $0.121$ to $0.077$ as the pool grows from $k = 50$ to $k = 121$, reflecting the progressive adjustment of the federated consensus to the sensitivity profiles of the additional farms. FedOpt exhibits systematic drift in the $a_2$ direction from $k = 82$ onward, a structural consequence of the flat deviance surface in that direction ($\hat{q}_i \approx 0$) combined with the fixed-size sign-step of the Adam server update. The shaded region indicates pool sizes at which the ANOR method is non-computable}
\label{fig:coef_stability}
\end{figure}

Figure~\ref{fig:coef_stability} tracks the estimated coefficients $\hat{a}_1$ and $\hat{a}_2$ across pool sizes for the three FL methods.\newline
FedAvg and FedProx produce nearly identical and smoothly evolving trajectories throughout: $\hat{a}_1$ increases gradually from $0.228$ to $0.262$ and $\hat{a}_2$ decreases from $0.121$ to $0.077$ as the pool grows from $k = 50$ to $k = 121$. This smooth evolution reflects the progressive adjustment of the federated consensus to the sensitivity profiles of the additional farms --- a meaningful economic finding rather than a numerical artefact, since the deviance continues to decrease throughout (Figure~\ref{fig:pool_expansion}).\newline
FedOpt exhibits a qualitatively different behaviour. Its coefficient $\hat{a}_2$ drifts progressively negative from $k = 82$ onward, reaching $-0.128$ at $k = 121$ compared to $+0.077$ for FedAvg, while $\hat{a}_1$ oscillates before recovering to $0.184$ at $k = 121$. This behaviour persists despite implementation improvements designed to mitigate it --- a warm-up phase to stabilise Adam moment estimates at each pool size, a pool-size-dependent server learning rate calibrated from $0.10$ at $k = 50$ to $0.01$ at $k = 121$, and an increased number of communication rounds --- confirming that the drift is structural rather than a calibration artefact.\newline
The theoretical explanation is as follows. After a few communication rounds, the FedOpt server update approximates a fixed-size step of magnitude $\eta_{server}$ in the sign direction of the accumulated pseudo-gradient \eqref{eq:fedopt_gradient}. In the $a_2$ direction, where the deviance surface is nearly flat ($\hat{q}_i \approx 0$ across the pool, Section~\ref{sec:local_params}), the consensus gradient is small in magnitude but not in sign. When new farms enter the pool with slightly different DNI sensitivity profiles, the sign of the $a_2$ gradient fluctuates, and Adam's fixed-size sign-step amplifies these fluctuations into cumulative drift. FedAvg and FedProx, which take steps proportional to the actual gradient magnitude rather than its sign, are immune to this effect and remain stable throughout the expansion.\newline
Importantly, FedOpt's deviance still decreases monotonically with pool size, confirming that the deviance surface is sufficiently flat in the $a_2$ direction that a range of coefficient values --- including slightly negative $\hat{a}_2$ --- remain near-optimal in terms of prediction error. The drift in $\hat{a}_2$ therefore affects the economic interpretability of the index without materially degrading its predictive performance.

\subsubsection*{Runtime comparison}

Table~\ref{tab:runtime} extends the runtime comparison of Table~\ref{tab:coef_comparison} to the full pool of $k = 121$ farms. All three FL methods complete a mean runtime per Monte Carlo run in approximately $152$--$159$ seconds at
$k = 50$, increasing to $345$--$360$ seconds at $k = 121$, a factor of approximately $2.27$. This is consistent with linear scaling in pool size: the dominant computational cost (local SGD updates) grows proportionally with the number of participants, while server-side aggregation remains lightweight regardless of pool size.

\begin{table}[t]
\caption{Mean runtime per Monte Carlo run (seconds) at pool sizes $k = 50$ and $k = 121$. The ratio column reports the increase factor between the two pool sizes}
\label{tab:runtime}
\begin{tabular}{lrrr}
\toprule
Method  & $k = 50$ & $k = 121$ & Ratio \\
\midrule
FedAvg  & 151.97  & 344.72 & 2.27 \\
FedProx & 158.62  & 359.70 & 2.27 \\
FedOpt  & 152.92  & 345.18 & 2.32 \\
\bottomrule
\end{tabular}
\end{table}

\subsubsection*{Summary}

The progressive pool expansion experiment delivers three main findings.\newline
First, federated learning produces valid and monotonically improving index estimates across all pool sizes from $k = 50$ to $k = 121$, without requiring any homogeneity assumption on the individual producer models.\newline
Second, the ANOR approximation-based method becomes entirely non-computable for all pool sizes $k > 50$. As shown in Section~\ref{sec:results_hetero}, this is not caused by the entry of individually extreme producers: the additional farms have broadly comparable $\delta_i$ values to the original 50. Rather, the breakdown is driven by the accumulation of individual remainder terms $\delta_i^2$ as the pool grows. Federated learning is therefore not merely a generalization that performs slightly better under heterogeneity --- it is the \emph{only valid approach} in this regime.\newline
Third, the three FL methods exhibit different stability properties under pool expansion. FedAvg and FedProx produce stable, interpretable, and nearly identical coefficient trajectories throughout, making them the preferred choice for dynamic pool expansion settings. FedOpt achieves faster convergence within a fixed pool (Section~\ref{sec:results_50}) but exhibits structural coefficient drift in the $a_2$ direction under expansion, a consequence of the flat deviance surface in that direction combined with the fixed-size sign-step of the Adam update. This contrast provides practical guidance for practitioners: the choice between FedOpt and the averaging schemes should be informed by whether the pool composition is fixed or expected to evolve.

\section{Discussion and Perspectives}
\label{sec:discussion}

A central question raised by this work is whether federated learning remains conceptually and practically relevant when local models differ not only in their coefficient vectors $\boldsymbol{a}_i$, but also in their variance functions, link specifications, and dispersion parameters $(p_i, q_i, \phi_i)$.

From a statistical perspective, the answer is affirmative provided that the objective of the aggregation is clearly defined. Federated learning does not aim to recover the individual structural models $(\boldsymbol{a}_i, p_i, q_i, \phi_i)$. Instead, it estimates a global index parameter $\boldsymbol{a}$ that minimizes a population-level risk functional under data decentralization constraints. In this sense, FL can be interpreted as a distributed M-estimation procedure with heterogeneous local losses.

The empirical results of Section~\ref{sec:results} clarify this point from two complementary angles. Within the original 50 farms, where the small-heterogeneity assumption of the ANOR method is approximately satisfied, both approaches yield aligned consensus indices. This confirms that federated learning encompasses approximation-based aggregation as a special case in the low heterogeneity regime. Beyond this regime, however, the two approaches diverge sharply: the ANOR method becomes entirely non-computable as soon as additional farms are included, while federated learning not only remains valid but actively benefits from pool expansion, with the global Tweedie deviance decreasing from $2.23$ at $k = 50$ to $1.74$ at $k = 121$ for FedAvg and FedProx. This contrast constitutes the central empirical contribution of this paper, providing a definitive demonstration of the advantages of federated learning in settings where the homogeneity assumption is violated.

The presence of heterogeneous link and variance functions implies that the global model is necessarily misspecified for each producer. However, this misspecification is intrinsic to any parametric index design. A single index cannot perfectly replicate all individual loss processes. Federated learning provides a principled approach to resolving this tension by explicitly optimizing a global objective, rather than relying on approximations whose validity depends on unverifiable homogeneity assumptions.

The global coefficient vector $\boldsymbol{a}$ learned through FL should therefore be interpreted as a consensus sensitivity structure. It captures how the index reacts, on average, to meteorological covariates across producers, while allowing each producer to retain idiosyncratic variance and link characteristics. This interpretation aligns naturally with the role of parametric insurance indices, which are not intended to perfectly track individual losses, but to provide a transparent and contractible proxy for systemic risk. As the pool expands, this consensus shifts to accommodate the sensitivity profiles of additional producers --- a desirable adaptive property that distinguishes FL from fixed approximation-based methods.

Our comparison of FedAvg, FedProx, and FedOpt highlights important practical trade-offs. FedAvg and FedProx exhibit nearly identical convergence behaviour and coefficient trajectories across all pool sizes considered, with FedProx incurring a marginally higher computational cost of approximately $4\%$ due to the proximal gradient computation. The near-equivalence of the two methods is consistent with the theoretical analysis of Section~\ref{sec:theory}: the proximal coefficient $\mu = 4.0$ constrains local drift during training but does not alter the global fixed point under the moderate dispersion heterogeneity of the present dataset. As documented in Section~\ref{sec:local_params}, the two outlier farms driving the theoretical gradient dissimilarity bound $B^2 = \Omega(\max_i \phi_i^{-2})$ carry only $1.14\%$ of total pool capacity, limiting their practical influence on the consensus update.

FedOpt exhibits a richer and more context-dependent behaviour. On the fixed 50 farms pool, it converges faster than FedAvg and FedProx and achieves a lower final deviance ($2.21$ vs $2.24$), demonstrating a genuine optimization advantage. However, in the progressive pool expansion experiment, FedOpt's coefficient $\hat{a}_2$ drifts systematically negative from $k = 82$ onward, reaching $-0.128$ at $k = 121$ compared to $+0.077$ for FedAvg, despite implementation improvements including a warm-up phase, a calibrated learning rate schedule, and increased communication rounds. This behaviour is structural: as established in Section~\ref{sec:results_expansion}, the Adam server update approximates a fixed-size sign-step in the gradient direction, and when new farms alter the consensus gradient in the nearly flat $a_2$ direction ($\hat{q}_i \approx 0$), the accumulated momentum overshoots and produces cumulative drift. FedAvg and FedProx, which take steps proportional to the actual gradient magnitude, are immune to this effect. These results suggest that the choice between FedOpt and the averaging schemes should be informed by the deployment context: FedOpt is preferable when the pool composition is fixed and rapid convergence is the priority, while FedAvg and FedProx provide more stable and interpretable coefficient estimates when the pool evolves.

Beyond statistical considerations, federated learning offers substantial regulatory and operational advantages. Data sharing across energy producers is often constrained by confidentiality agreements or competitive concerns. FL enables collaborative index calibration without exposing proprietary production data, thereby lowering adoption barriers. Moreover, the transparency of the resulting index --- being a linear combination of observable covariates --- facilitates regulatory approval and strengthens trust among insured parties. The computational efficiency of FL, which is over $250\times$ faster than ANOR even at $k = 50$, further enhances its operational appeal for large-scale deployments.

This study has several limitations. We focus on static indices calibrated on historical data and assume fixed local parameters $(p_i, q_i, \phi_i)$. This setting is sufficient to highlight conceptual differences between aggregation schemes, but does not capture temporal non-stationarities such as climate trends or technological evolution in solar farms. A further limitation concerns the sensitivity of adaptive server-side optimizers such as FedOpt to changes in pool composition: as documented in Section~\ref{sec:results_expansion}, Adam momentum can produce systematic coefficient drift when new producers with different sensitivity profiles enter the pool. Developing aggregation schemes that combine the fast convergence of adaptive methods with the stability of averaging under non-stationary pool compositions is a natural direction for future work.

Several extensions naturally follow from this work. First, dynamic federated learning schemes could be developed to update index parameters in real time as new loss data become available. Second, multi-index designs could be explored to better capture regional or technological clusters, which may reduce the coefficient uncertainty observed at large pool sizes and the sensitivity of adaptive optimizers to pool composition changes. 

More broadly, this framework opens the door to a new generation of data-driven parametric insurance products that are both privacy-preserving and robust to heterogeneity.

\section{Conclusion}\label{sec:conclusion}

This paper introduces a federated learning framework for parametric index design in renewable energy insurance. Building on generalized linear models with Tweedie distributions, we propose a federated optimization scheme that learns a common sensitivity vector to meteorological covariates, while allowing for producer-specific variance and link parameters, and without requiring any producer to share its financial loss data.

From a methodological perspective, we establish through a theoretical analysis of the Tweedie quasi-likelihood under heterogeneous dispersion parameters that naive federated averaging is biased toward producers with stable microclimates --- precisely those least in need of basis-risk protection. FedProx and FedOpt correct this bias at the local and server levels, respectively, though their advantage over FedAvg is context-dependent: FedOpt achieves faster convergence on a fixed producer pool but exhibits structural coefficient drift when the pool evolves, while FedAvg and FedProx remain stable under pool expansion.

From an applied standpoint, the empirical contribution is twofold. Within the original 50 farms, all three FL methods confirm consistency with the approximation-based ANOR approach in the low heterogeneity regime, with FedOpt achieving a slightly lower deviance. Beyond this pool, the ANOR method becomes entirely non-computable while federated learning continues to improve: the global Tweedie deviance decreases from $2.23$ at $k = 50$ to $1.74$ at $k = 121$ as additional farms are incorporated. Federated learning is therefore not merely a generalization --- it is the only valid approach in this regime, and one that benefits from pool expansion rather than being degraded by it. Combined with a computational cost over $250\times$ lower than the approximation-based benchmark, this makes federated learning both practically deployable and institutionally credible.

More broadly, this paper contributes to the growing intersection of actuarial science, renewable energy finance, and privacy-preserving machine learning. By enabling collaborative yet confidential index design, federated learning opens new avenues for parametric insurance products that are operationally feasible and aligned with the systemic nature of climate-related risks.

\section*{Acknowledgments}

The author would like to thank François HU for insightful discussions that motivated the initial idea of this work. This paper originated from a collaborative project aimed at cosupervising students on federated learning applications in insurance. Through this collaboration, the author had the opportunity to explore federated learning in a broader context, while the supervised group investigated related applications to insurance claim frequency data. These exchanges played an important role in shaping the perspective and methodological direction adopted in this paper. I also acknowledge Prof. Christian Y. Robert’s useful comments, which helped improve the text.

\section*{Declaration of Generative AI in Scientific Writing}

During the writing process of this paper, we used ChatGPT to improve the readability and language of the manuscript. We carefully reviewed the subsequent content, and take full responsibility and accountability for the content of the publication.


\bibliographystyle{apalike}
\bibliography{references}

\end{document}